\documentclass[10pt,twocolumn,letterpaper]{article}
\usepackage[table]{xcolor}
\usepackage{booktabs}
\usepackage{iccv}
\usepackage{times}
\usepackage{epsfig}
\usepackage{graphicx}
\usepackage{algorithm}
\usepackage{algorithmic}
\usepackage{amsmath}
\usepackage{amssymb}
\usepackage{multirow}
\usepackage{multicol}
\usepackage{makecell}
\usepackage{threeparttable}
\usepackage{subcaption}

\def\ie{\textit{i.e.}}
\newcommand{\best}[1]{\textcolor{red}{#1}}
\newcommand{\second}[1]{\textcolor{blue}{#1}}
\newcommand{\D}[2]{\mathcal{D}_{KL}(#1||#2)}
\newcommand{\E}[2]{\mathbb{E}_{#1}[#2]}
\def\p{p_\Theta}
\def\P{\mathcal{P}}
\def\F{F_{\Theta}}

\newcommand{\para}[1]{\noindent\textbf{#1}}

\usepackage[pagebackref=true,breaklinks=true,letterpaper=true,colorlinks,bookmarks=false]{hyperref}
\newcommand{\wh}[1]{\textcolor{black}{#1}}
\newcommand{\bihan}[1]{\textcolor{black}{#1}} 
\iccvfinalcopy 

\def\iccvPaperID{2118} 

\ificcvfinal\fi

\begin{document}

\title{ExposureDiffusion: Learning to Expose for Low-light Image Enhancement}
\author{Yufei Wang$^1$, Yi Yu$^1$, Wenhan Yang$^2$, Lanqing Guo$^1$, Lap-Pui Chau$^3$, Alex C. Kot$^1$, Bihan Wen$^1\thanks{Corresponding author.}$\\
$^1$Nanyang Technological University \quad $^2$Peng Cheng Laboratory \\
$^3$The Hong Kong Polytechnic University \\
{\tt\small \{yufei001, yuyi0010, lanqing001, eackot, bihan.wen\}@ntu.edu.sg}\\ {\tt\small yangwh@pcl.ac.cn \quad lap-pui.chau@polyu.edu.hk}
}

\maketitle
\ificcvfinal\thispagestyle{empty}\fi

\begin{abstract}
\vspace{-0.1cm}
%
\wh{Previous raw image-based low-light image enhancement methods predominantly relied on feed-forward neural networks to learn deterministic mappings from low-light to normally-exposed images. However, they failed to capture critical distribution information, leading to visually undesirable results.}
\wh{This work addresses the issue by seamlessly integrating a diffusion model with a physics-based exposure model.
%
Different from a vanilla diffusion model that has to perform Gaussian denoising, with the injected physics-based exposure model, our restoration process can directly start from a noisy image instead of pure noise.
As such, our method obtains significantly improved performance and reduced inference time compared with vanilla diffusion models.
%
To make full use of the advantages of different intermediate steps, we further propose an adaptive residual layer that effectively screens out the side-effect in the iterative refinement when the intermediate results have been already well-exposed.}
\bihan{The proposed framework can work with both real-paired datasets, SOTA noise models, and different backbone networks.}
%
\bihan{We evaluate the proposed method on various public benchmarks, achieving promising results with consistent improvements using different exposure models and backbones.}
%
\wh{Besides, the proposed method achieves better generalization capacity for unseen amplifying ratios and better performance than a larger feedforward neural model when few parameters are adopted.} The code is released at 
\url{https://github.com/wyf0912/ExposureDiffusion}.
\end{abstract}

\vspace{-0.3cm}
\section{Introduction}
Over the past few years, learning-based methods for low-light image enhancement~\cite{liu2021benchmarking, li2021low, jin2023enhancing} have gained significant attention and made remarkable progress, and most of them are conducted in the sRGB space.
%
%
%
\begin{figure}[t]
    \centering
    \includegraphics[trim=10 0 0 0, clip, width=1\linewidth]{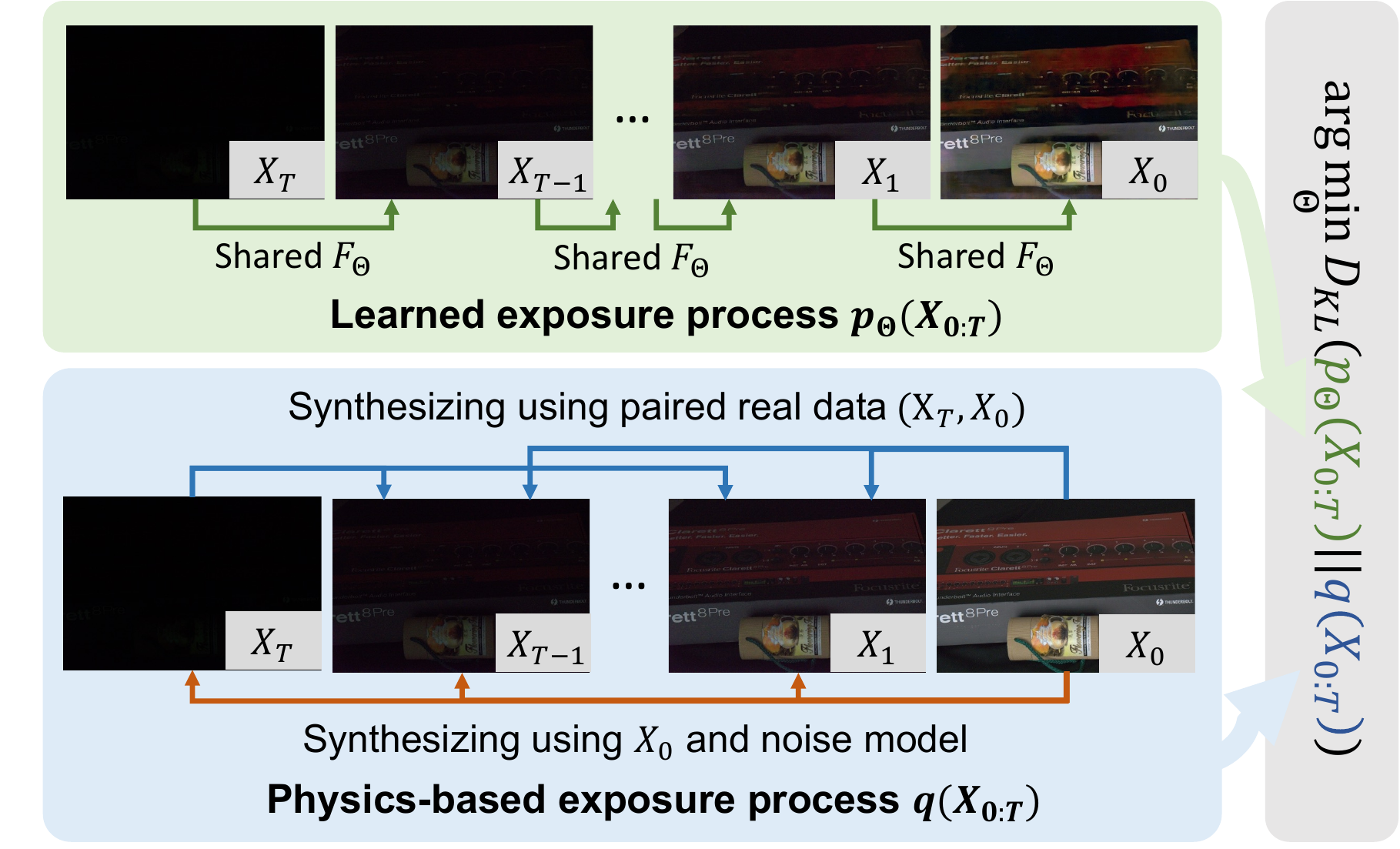} 
    \vspace{-0.65cm}
    \caption{
    We propose to simulate the physics-based exposure process using a shared neural network $F_\Theta$ \wh{in a progressive manner}.
    The learned exposure process is optimized to \wh{approximate} the physics-based exposure process by minimizing the derived \wh{variational} upper bound of the KL divergence \wh{of their distributions}.
    Besides, the proposed strategy can be applied on \wh{real-captured} paired data (blue array) \wh{and} synthetic data with different noise models (orange array), and \wh{in} different backbone networks. 
    Benefiting from learning a continuous exposure process, the proposed method can be applied to \wh{work with} an arbitrary amplifying factor, and better performance can be achieved by the iterative refinement \wh{process}.}
\vspace{-0.5cm}
\end{figure}
Recently, the enhancement in the raw space is demonstrated to have unique advantages over sRGB spaces~\cite{huang2022towards}. For example, raw images provide a higher dynamic range, leading to better performance in extremely dark environments. Besides, the linear correlation between the low-light and normally-exposed images \wh{prevents improper exposure level adjustment in the enhancement process.}
In addition, the noise modeling in the raw space is \wh{more straightforward} than that in the sRGB space by ruling out the effect of increasingly complicated image signal processing pipelines. \wh{In such space, the domain gap between synthetic and captured data is small and} the model trained with paired synthetic images exhibits comparable or even better performance than that of real-captured data~\cite{wei2020physics, feng2022learnability}. 
While promising progress is achieved, \wh{the prevailing approach remains to} learn a deterministic mapping based on \wh{feedforward neural networks}. For the images captured in extremely dark environments, \wh{this one-step} enhancement/denoising \wh{process}\footnote{\wh{As exposure changes can be approximated with a linear transform in the raw space, low-light image enhancement in the raw image space is regarded as a denoising task in most previous works.}} \wh{fails to characterize the distribution information and}
usually obtain \wh{undesirable results}. \wh{For example,} there may still exist some residual \wh{noise.}
Besides, \wh{existing} works mainly pay \wh{attention to more accurate noise modeling}.
\wh{The work of effectively incorporating the noise model in the raw space into a learnable model for improved enhancement remains unexplored.}

Most recently, generative model-based image restoration methods~\cite{lugmayr2020srflow, wang2021low, saharia2022image} exhibit appealing performance and \wh{pleasing} perceptual quality in image restoration tasks. 
\wh{Among these generative models}, diffusion models~\cite{song2020denoising, ho2020denoising}~\wh{stand out for their capacity to model} a complicated distribution \wh{with} arbitrary neural networks in a \wh{progressive} manner and exhibit great success in image generation and restoration tasks \cite{saharia2022image}.
\wh{Different types of forward processes have been explored in diffusion models}, \eg, the unified framework for the Wiener process~\cite{zhang2022gddim}.
\wh{Nonetheless, they are inadequate to accurately simulate real exposure processes.}
First, the low-light image is naturally not an intermediate step of the \wh{vanilla} diffusion process. Therefore, the reverse (denoising) process needs to start from \wh{pure} noise and involves a relatively large number of inference steps, \wh{which hinders the real applications.}
Second, since the \wh{vanilla} diffusion models need to have the capacity of removing Gaussian noise with different noise levels, it usually requires extra model capacity compared with feedforward \wh{neural} networks.

To \wh{address} aforementioned \wh{issues, we propose a novel approach to effectively inject noise models in raw space into an end-to-end learnable progressive model}, named \textit{ExposureDiffusion}. 
Specifically, we propose to simulate the exposure process using a \wh{progressive} shared network \wh{to minimize} the divergence between the simulated process and the real one by \wh{optimizing the proposed variational} upper bound.
Since \wh{the intermediate steps of the progressive process} all obey the physics-based noise distribution, the restoration process can \wh{directly} start from a noisy image instead of pure noise.
\wh{This design significantly benefits the low-light enhancement/denoising in two dimensions.
First, the proposed method no longer requires removing Gaussian noises and only needs to learn the process of real-noise denoising, leading to smaller requirements of model capacity.
%
Second, the proposed method greatly reduces the required number of inference steps, which has the potential to significantly benefit real applications.}
Besides, we further propose an adaptive residual layer to \wh{dynamically fuse} different denoising strategies for the areas with different noise-to-signal ratios.
\wh{This strategy effectively screens out the side-effect in the iterative refinement when the intermediate results have been already well-exposed.}
The proposed method can be \wh{applied to} both paired real-captured data, synthetic data with different noise models, and different backbone networks. 
Experimental results demonstrate that the proposed method can achieve \wh{significant} improvement \wh{jointly with both real/synthetic} exposure process and backbone networks. The proposed method, which employs the noisy-to-fine strategy, also exhibits superior generalization capability. Our main contributions are summarized as follows:
\begin{itemize}
\vspace{-0.2cm}
    \setlength\topsep{-1cm}
    \setlength\itemsep{-0.10cm}
    \item We propose the first diffusion-based model for low-light image enhancement in the raw image space. The modeling of the process is \wh{inspired and constructed strictly according to the physical noise model}.
    \wh{This design enables restoration from any intermediate step of the diffusion process and eliminates the need for the Gaussian denoising process. As a result, the available model capacity and inference efficiency are significantly improved.}
    \item We further propose an adaptive residual layer to \wh{dynamically} adopt different denoising strategies for areas with different noise-to-signal ratios.
    \wh{This strategy effectively screens out the side-effect in the iterative refinement when the intermediate results have been already well-exposed.}
    \item Extensive experimental results on two public datasets demonstrate the \wh{significant} performance improvement of the proposed method combined with \wh{state-of-the-art} noise models/backbones. Besides, the proposed method exhibits better generalization \wh{capacity} compared with feedforward \wh{neural} networks and possesses fewer parameters and faster speed \wh{to achieve competitive} performance. 
\end{itemize}
\vspace{-0.4cm}
\begin{figure*}[t]
    \centering
    \vspace{-0.3cm}
    \includegraphics[width=1\linewidth, trim=0 0 0 0, clip]{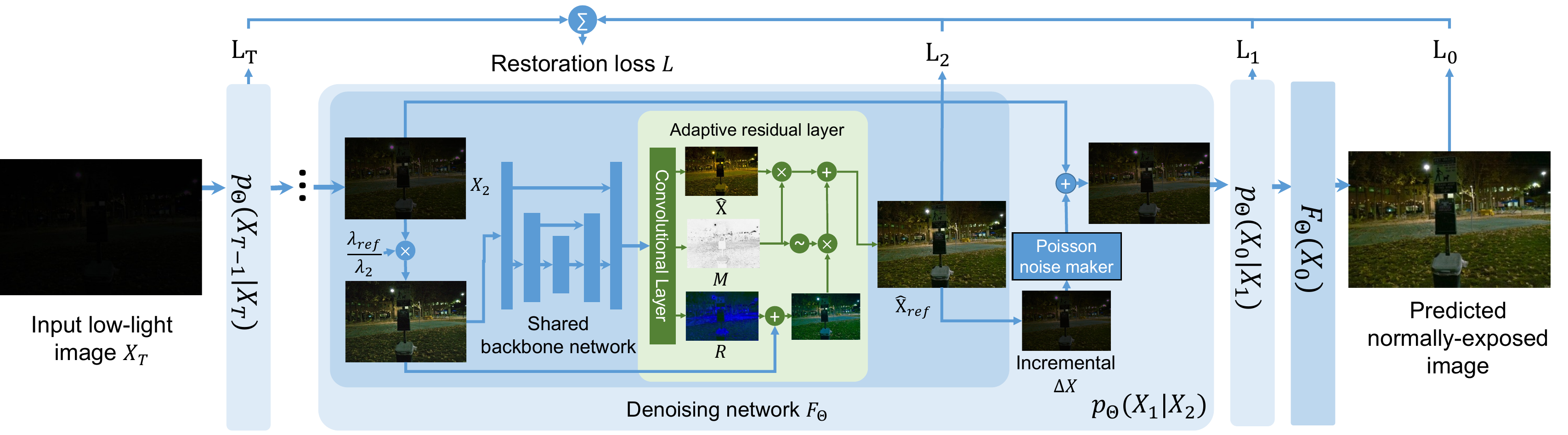} 
    \vspace{-0.2cm}
    \caption{The proposed method follows an overall framework where the final results are achieved through a process of progressive refinement. The adaptive residual layer (in green) can be combined with any backbone network. For both training and inference, the process starts from a low-light image $X_T$, and images with longer exposure time are gradually achieved. The image reconstruction losses $L_t$ of different steps are all used for training, and $F_\theta(X_0)$ is the final result.}
    \label{fig:framework}
    \vspace{-0.5cm}
\end{figure*}

\section{Related works}
\para{Low-light image enhancements.} A great number of low-light image enhancement methods based on deep learning have been proposed in the past few years~\cite{zhang2021rellie, huang2022deep, wang2022ultra, guo2022low, xu2022snr, jin2022unsupervised, jin2023enhancing}. The mainstream of these methods is based on supervised learning, \ie, training a mapping from low-light images to normally-exposed images. For example, LLNet \cite{lore2017llnet} proposes an autoencoder to enhance the visibility of low-light images. To obtain better perceptual quality, \cite{shen2017msr, tao2017llcnn, lv2018mbllen, ren2019low} propose to utilize multi-scale features to better learn the global content and salient structures. Retinex theory is also widely used as the prior knowledge to guide the disentanglement of reflection and illumination maps \cite{wei2018deep, wang2019progressive, wei2018deep, guo2022enhancing}. Unfolding/unrolling-based methods \cite{liu2021ruas, zheng2021adaptive, wu2022uretinex} are explored to better utilize priors of low-light image enhancement. Recently, the explicit modeling of the conditional distribution of normally-exposed images is explored in \cite{wang2021low}, showing superior perceptual quality. Besides the aforementioned enhancement methods in sRGB space, the enhancement in the image space gradually attracts increasing attention recently~\cite{huang2022towards, chen2018learning, punnappurath2022day, dong2022abandoning} due to its unique advantages. Current research in the raw image space mainly focuses on the realism of noise modeling \cite{wei2020physics, zhang2021rethinking, feng2022learnability, jiniccv23led, wang2023raw, wang2023beyond} so that the synthetic training data can have a smaller domain gap. Specifically, Poisson-Gaussian \cite{foi2008practical} is a basic and widely used noise model, which assumes the noises include signal-independent Gaussian noise and signal-dependent Poisson noise. The following works mainly improve the modeling of the signal-independent noise, \eg, additionally modeling the row noise \cite{wei2020physics} and the dark shading~\cite{feng2022learnability}. However, the exploration of training strategies to better utilize the clear formulation of the degradation process in the raw space is still lacking.

\para{Diffusion models.}
Recently, diffusion-based~\cite{song2020denoising, ho2020denoising} image restoration models \cite{whang2022deblurring, wang2021s3rp} exhibit remarkable performance by using the degraded image as the conditional input. For example, \cite{chung2022improving} proposes a diffusion model for different inverse problems by using manifold constraints. By altering the reverse diffusion process, \cite{lugmayr2022repaint} exhibits good performance in free-form inpainting. \cite{kawardenoising} utilizes the pre-trained diffusion models to conduct multiple restoration tasks, \eg, super-resolution, and deblurring. \cite{saharia2022image} proposes a conditioned diffusion model for super-resolution that uses the low-resolution image as a part of the input, and learns the whole process in an end-to-end manner. Besides, \cite{saharia2022palette} proposes to handle different image-to-image tasks based on conditional diffusion models, and diffusion models are used as plug-and-play image priors in~\cite{graikosdiffusion}. While promising results are achieved, they mainly focus on super-resolution, deblurring, inpainting, and colorization, for which the synthetic dataset is relatively easy to be obtained. Diffusion models, especially the physics-based model, for low-light image enhancement are still left to be explored. Besides, some efforts \cite{zhang2022gddim, pokledeep, guo2022shadowdiffusion, ma2022accelerating} aim to speed up the sampling process of diffusion models. For example, \cite{zhang2022gddim} speeds up the inference speed by reducing the number of sampling steps and \cite{pokledeep} uses deep nonequilibrium approaches to find the results after convergence. However, the initial states are still usually from pure noise even for the conditional image restoration task~\cite{guo2022shadowdiffusion, saharia2022image}. 

\begin{table*}[tbp]
    \centering
    \vspace{-0.1cm}
    \scalebox{0.95}{
    \begin{threeparttable}
    \begin{tabular}{cccc}
    \toprule
         & Unconditional diffusion \cite{song2020denoising, ho2020denoising} & Conditional diffusion \cite{saharia2022image} & Exposure diffusion (Ours) \\
    \midrule
     Objective  & maximizing $\p(X)$ & maximizing $\p(X|Y)$ & \makecell[c]{minimizing KL divergence with\\the real exposure process}  \\
     Initial state $X_T$ & $X_T \sim \mathcal{N}(0, 1)$ & $X_T \sim \mathcal{N}(0, 1)$ & $X_T \sim q(X_T)$ \tnote{1}\\
     Assumption &  \makecell[c]{$q(X_{t}|X_{t-1}):=$\\$\mathcal{N}(X_t;\sqrt{1-\beta_t}X_{t-1}, \beta_t \mathbf{I})$} & \makecell[c]{$q(X_t|X_{t-1}):=$\\$\mathcal{N}(X_t;\sqrt{1-\beta_t}X_{t-1}, \beta_t \mathbf{I})$} & \makecell[c]{$q(X_{t-1}|X_t, X_{ref}):=$\\$\P (\frac{X_{t-1}-X_{t}}{K};\frac{(\lambda_{t-1}-\lambda_{t})X_{ref}}{\lambda_{ref}K})$} \tnote{2}\\
     Reverse process & \makecell[c]{$\p(X_{t-1}|X_t):=$\\$\mathcal{N}(X_{t-1};\mu_\Theta(X_t, t), \sigma_t^2 \mathbf{I})$} 
     & \makecell[c]{$\p(X_{t-1}|X_t, Y):=$\\$\mathcal{N}(X_{t-1};\mu_\Theta(X_t, Y, t), \sigma_t^2 \mathbf{I})$} 
     & \makecell[c]{$\p(X_{t-1}|X_{t}):=$\\$\P (\frac{X_{t-1}-X_{t}}{K};\frac{(\lambda_{t-1}-\lambda_{t})\F(X_t)}{\lambda_{ref}K})$} \\
     Training & The expectation over $q(X_t|X_0)$ & The expectation over $q(X_t|X_0)$ & The expectation over $\p(X_{t}|X_T)$ \tnote{3}\\
    \bottomrule
    \end{tabular}
    \vspace{-0.1cm}
      \begin{tablenotes}[para,flushleft]
      \item[1] We start from a low-light image instead of a pure noise sampled from a Gaussian distribution. \\
      \item[2] The Markov process is formulated based on the physics-based model of the exposure process. \\
      \item[3] The cumulative error caused by progressive refinement is alleviated by minimizing the loss $L_t$ over the expectation of $p_\theta$.
      \end{tablenotes}
    \end{threeparttable}}
    \vspace{-0.25cm}
    \caption{A comparison between the proposed algorithm and vanilla diffusion models.}
    \label{tab:diff}
    \vspace{-0.55cm}
\end{table*}

\section{Methodology}
\vspace{-0.1cm}
\subsection{Preliminary}
\vspace{-0.1cm}
A raw image $X_t$ can be formulated as follows
\begin{equation}
    X_t = \lambda_t K I + N,
\label{eq:raw_img}
\end{equation}
where $\lambda_t$ represents the exposure time, $K$ is the overall system gain, $I$ is the rate of the photoelectrons which is proportional to the scene irradiation, and $N$ is the summation of all noise sources. 
The formulation of the noise summation $N$ in the raw image can be simplified as follow
\begin{equation}
    N = K N_p + N_{ind},
\label{eq:noise_model}
\end{equation}
where $N_p$ is the photon shot noise, and $N_{ind}$ is the signal independent noise. The photon shot noise obeys a Poisson distribution as follows
\begin{equation}
    (\lambda_t I+N_p) \sim \mathcal{P}(\lambda_t I).
\end{equation}
The target of low-light image enhancement is to predict the normally-exposed image $X_{0}$ given an image $X_T$ with a short exposure time, \ie, $\lambda_{0}>\lambda_{T}$ \footnote{We assume that $\lambda_t$ decreases with the increase of $t$. Details are in the supplementary material.}. 
The mainstream of the previous works aims to minimize the reconstruction loss between the restored image and the reference image by optimizing a deep network parameterized by $\Theta$ as follows
\begin{equation}
    \Theta = \arg \min_\Theta \mathcal{L}(F_{\Theta} (X_T), X_{0}),
\end{equation}
where $\mathcal{L}$ can be any pixel-wised reconstruction loss, \eg, $L1$ and $L2$ loss. 
However, due to the over-simplified assumption of the distribution of normally-exposed images, such a training paradigm usually leads to the residual of unnatural artifacts in outputs \cite{wang2021low}. %
To integrate the advantages of the learnable conditional distribution of normally-exposed images and the clear formulation of the noise in raw space simultaneously, we propose a novel method to learn to expose which is illustrated in the following section.
%


\subsection{Learning to expose}
\vspace{-0.05cm}
\subsubsection{The formulation of the training objective}
\vspace{-0.05cm}
To enhance the visibility of a low-light image $X_T$, we aim to learn a model $F_\Theta$ that can maximize the likelihood of its reference images $X_{0}$ parameterized by $\Theta$ over the distribution of the training data $q(X_0, X_T)$, \ie,
\begin{equation}
    \Theta = \arg \max_\Theta \E{q}{p_{\Theta}(X_{0}|X_T)}.
\end{equation}
Due to the difficulty of directly estimating the likelihood of an image, we further formulate $p_{\Theta}(X_{0}|X_T)$ as follows,
\begin{equation}
\small
    p_{\Theta}(X_{0}|X_T) = \frac{\int p_{\Theta} (X_{0:T}) d X_{1:T-1}}{p_{\Theta}(X_T)},
\end{equation}
so that maximizing $\E{q}{\log[p_{\Theta}(X_{0}|X_T)]}$ is equivalent to maximizing the likelihood of the joint distribution $\E{q}{\log[\p(X_{0:T})]}$, \ie, minimizing the cross entropy between the learned exposure process and the real one. Therefore, we propose to minimize the upper bound of the divergence between $p_\Theta(X_{0:T})$ and $q(X_{0:T})$ as follows
\begin{equation}
\small
\begin{aligned}
    &\D{\p(X_{0:T})}{q(X_{0:T})} \leq 
    \\
    &\E{q(X_{ref})}{\D{\p(X_T)}{q(X_T|X_{ref})} + \\
    &\sum_{t=1}^{T} \E{\p(X_t)}{\D{\p(X_{t-1}|X_{t})}{q(X_{t-1}|X_t,X_{ref})}}},
\end{aligned}
\label{eq:bound}
\end{equation}
where $X_{ref}$ is the expected clean image, \ie, when $N$ in Eq.~\ref{eq:raw_img} is a zero matrix, and $q(X_{0:T})$ is the ground-truth distribution of the exposure process. $X_0$ can be approximately regarded as $X_{ref}$ if its exposure time is long enough. The detailed derivation can be found in the supplement. 
\subsubsection{Training strategy}
We do not need to optimize the first term in the proposed upper bound in Eq. \ref{eq:bound} since we aim to learn a restoration model instead of a generative model.
For the second term in Eq. \ref{eq:bound}, it calculates the divergence between the distribution of predicted image $p_{\Theta}(X_{t-1}|X_{t})$, \ie, the image with slightly longer exposure time and higher signal-noise-ratio (SNR) than $X_{t}$, and the real exposure process. The real exposure process $q(X_{t-1}|X_t,X_{ref})$ is well-defined based on the noise model in Eq. \ref{eq:noise_model} as follows
\begin{equation}
\small
    q(X_{t-1}|X_t,X_{ref}) = \P \left( \frac{X_{t-1}-X_{t}}{K};\frac{(\lambda_{t-1}-\lambda_{t})X_{ref}}{\lambda_{ref}K}\right) ,
\end{equation}
where $\frac{(\lambda_{t-1}-\lambda_{t})X_{ref}}{\lambda_{ref}K}$ is the rate of the Poisson distribution $\P$. Namely, the increment part in the count of photons obeys Poisson distribution. For the design of $\p(X_{t-1}|X_{t})$, the increment part is assumed to obey the following Poisson distribution
\begin{equation}
\small
    \p(X_{t-1}|X_t) = \P \left(\frac{X_{t-1}-X_{t}}{K};\frac{(\lambda_{t-1}-\lambda_{t})\F(X_t)}{\lambda_{ref}K}\right),
\end{equation}
which can be trivially sampled as follows,
\begin{equation}
\small
\begin{split}
    X_{t-1} = X_t + \Delta X + K \Delta N_p, \\
    \text{where } \Delta X= \frac{(\lambda_{t-1}-\lambda_t) \F(X_t)}{\lambda_{ref}} \\
    \text{and } \left(\frac{\Delta X}{K} + \Delta N_p \right) \sim \mathcal{P}\left(\frac{\Delta X}{K}\right).
\end{split}
\label{eq:sampling}
\end{equation}
After defining the formulation of two terms in $\D{\p(X_{t-1}|X_{t})}{q(X_{t-1}|X_t,X_{ref})}$, the pixel-wise reconstruction loss can be optimized as follows
\begin{equation}
\small
    L_t =\E{\p}{\F(X_t) \cdot \log \frac{\F(X_t)}{ X_{ref}}+X_{ref}-\F(X_t)},
\label{eq:lt}
\end{equation}
and the derivation can be found in the supplement. It is worth noting that the expectation of $L_t$ is calculated over the distribution of $\p$, \ie, the input image with shorter exposure time should be sampled from the distribution parameterized by $\Theta$. The details of the proposed training and inference procedures are in Algorithms \ref{alg:training}, \ref{alg:inference} and Fig.~\ref{fig:framework}.

\begin{figure}[t]
\vspace{-3mm}
\begin{algorithm}[H]
    \caption{Training (default $T = 2$)}
    \scalebox{0.9}{
    \begin{minipage}{1\linewidth}
    \begin{algorithmic}[1]
        \WHILE{not converged}
        \STATE sample $(X_T, \lambda_T, X_{ref}, \lambda_{ref})$ from either synthetic dataset or real-captured paired dataset
        \STATE $L = 0$
        \FOR {$t = T, T-1, ..., 0$}
        \STATE $\hat{X}_{ref} = \F(X_t)$
        \STATE $L=L+L_t(\hat{X}_{ref}, X_{ref})$ following Eq. \ref{eq:lt}
        \IF{$t>0$}
        \STATE $X_{t-1} \sim \p(X_{t-1}|X_t)$ following Eq. \ref{eq:sampling}.
        \ENDIF
        \ENDFOR
        \STATE Perform a gradient descent step on $\nabla_{\Theta} L $ 
        \ENDWHILE
    \end{algorithmic}
    \end{minipage}
    }
    \label{alg:training}
\end{algorithm}
\vspace{-1cm}
\end{figure}

\begin{figure}[t]
\begin{algorithm}[H]
    \caption{Inference (default $T = 1$)}
    \scalebox{0.9}{
    \begin{minipage}{1\linewidth}
    \begin{algorithmic}[1]
    \STATE \textbf{Input:} model $F_\theta$, desired exposure time $\lambda_{ref}$, low-light image $X_T$, and its exposure time $\lambda_T$
        \FOR {$t = T, T-1, ..., 0$}
        \STATE ${\hat{X}_{ref}} = \F(X_t)$
        \STATE $X_{t-1} \sim \p(X_{t-1}|X_t)$ following Eq. \ref{eq:sampling}.
        \ENDFOR
        \STATE return $F_\Theta(X_0)$
    \end{algorithmic}
    \end{minipage}
    }
    \label{alg:inference}
\end{algorithm}
\vspace{-0.9cm}
\end{figure}

\vspace{-0.1cm}
\subsubsection{Adaptive residual layer}
Although there are no inherent restrictions on the network design imposed by the proposed algorithm, making specific modifications to the network can further enhance its performance.
Specifically, we find that although the proposed inference algorithm can overall improve the quality of restored images, it may increase the error in bright areas, \eg, light bulbs. Namely, due to the high signal-noise ratio in the bright area, the reconstructed result of the initial step may be the most accurate one and may be further degraded by subsequent refinements. To solve this problem, we propose an adaptive residual layer. Specifically, the network $\F$ is designed to predict normally-exposed image $X_{ref}$, the noise residual $R=X_{ref}-\frac{\lambda_{ref}X_t}{\lambda_t}$, and a soft mask $M$ simultaneously, and the final output $\F(X_t)$ is as follows
\begin{equation}
    \F(X_t) = M \cdot \lfloor \hat{X} \rceil + (1-M) \cdot \lfloor\frac{\lambda_{ref}}{\lambda_t} X_t+\hat{R}\rceil, 
\label{eq:ARL}
\end{equation}
where $\hat{X}$ and $\hat{R}$ are the predicted reference image and the residual respectively, and $\lfloor \rceil$ is the $[0, 1]$ clip operation\footnote{To stabilize the training process, the input to the network is amplified to match the brightness of the reference images. However, before sampling $\p(X_{t-1}|X_t)$, the images are converted to photon counts.}. More specifically, the only change in the architecture is the increase in the number of output channels. For example, if the number of raw image channels is $4$, the proposed network will have $9$ channels for output, which include $4$ channels for $\hat{X}$, $4$ channels for $R$, and one channel for $M$.

\subsection{Comparison with diffusion models}
In this section, we compare the proposed algorithm with diffusion models \cite{song2020denoising, ho2020denoising, saharia2022image} since they all involve density estimation and progressive refinement. The differences are summarized in Table \ref{tab:diff}. As shown in the table, the proposed method has a different motivation and formulation, leading to different inductive biases and model performance. The main advantages of the proposed method are as follows: first, each intermediate step $X_t$ obeys the physics-based noise distribution in the proposed method while it is not satisfied in previous diffusion models. 
This consistency makes the proposed model a better generalization ability towards different noise levels and does not need to spend model capacity to learn the Gaussian denoising. 
In addition, fewer inference steps can be achieved by staring from a low-light image instead of pure noise.
%
%
Besides, explicitly integrating the accumulative error, \ie, the divergence between $q(X_{t-1}|X_T, X_{ref})$ and $\p(X_{t-1}|X_t)$, into the training process enables the proposed method to achieve higher fidelity results compared to vanilla diffusion models. More details can be found in the supplementary material.

\begin{figure*}[t]
    \centering
    \includegraphics[width=1\linewidth, clip, trim=0 0 0 10]{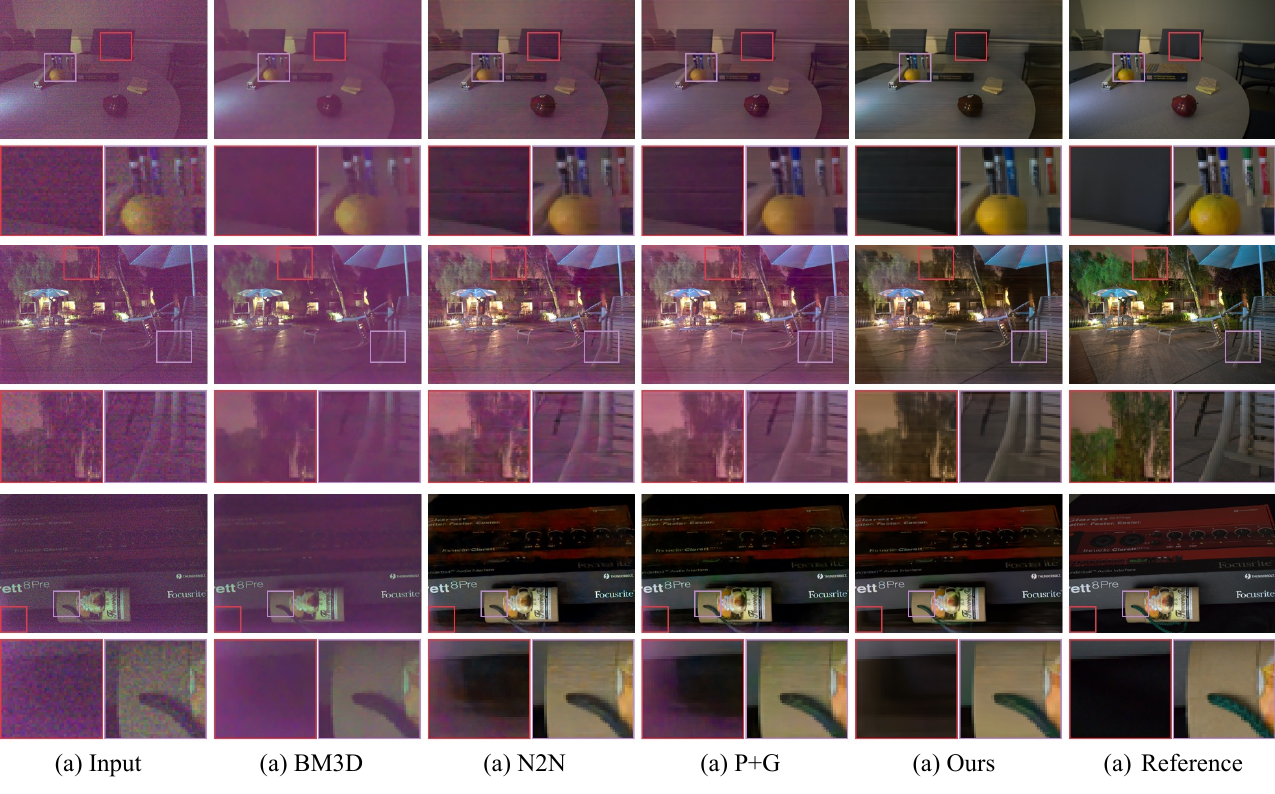} 
    \vspace{-0.45cm}
    \caption{Low-light image enhancement results on both indoor and outdoor environments after the same ISP pipeline for better visualization. The results in \textit{Ours} are obtained by using the same noise model and backbone as P+G. }
    \label{fig:vis_res}
\end{figure*}

\vspace{-0.1cm}
\section{Experiment}
\vspace{-0.1cm}
\subsection{Experimental setting}
\para{Implementation details.}
We evaluate the performance of the proposed method on two widely-used raw image low-light enhancement datasets: ELD~\cite{wei2020physics} and SID~\cite{chen2018learning}. Specifically, for the commonly used Bayer array, SID~\cite{chen2018learning} contains 2697 pairs of raw images under dark environments, which are captured under different ISO and amplification ratios, \eg, $\times100$, $\times250$, and $\times300$. We use the same split as~\cite{wei2020physics} for SID~\cite{chen2018learning} dataset, and train all the models using its training set. In this work, ELD~\cite{wei2020physics} is used for additional evaluation of the generalization ability of models under different scenarios and devices.
For the modeling of the real noise distribution, we adopt the widely-used P+G model \cite{foi2008practical, wei2020physics} as the baseline, in which the distribution of the signal-dependent noise is modeled as Poisson distribution, and the signal-independent noise is set to Gaussian noise as default. 
The impact of the choice of signal-independent noise is further explored in Sec. 
\ref{sec:noise_model}. For all experiments, we use the real-captured paired data for evaluation. More details can be found in the supplementary material.

\subsection{Comparison with different methods.} Since there are few works exploring the network design of the low-light image enhancement under raw space, we include the following competitors for comparison: the typical non-deep methods which do not require paired data for training, \eg, BM3D~\cite{dabov2007image}, and A-BM3D~\cite{makitalo2010optimal}. The model that utilizes paired noise image for training, \ie, Noise2Noise (N2N)~\cite{lehtinen2018noise2noise}. The model trained with the synthetic paired data, \ie, P+G \cite{foi2008practical, wei2020physics}. Specifically, P+G \cite{foi2008practical, wei2020physics} means we train the model using the synthetic image with the Possion-Gaussian noise model. Ours uses the same settings with P+G \cite{foi2008practical, wei2020physics}, \eg, the same noise model, and almost the same network architecture (the only difference is the number of input/output channels), except that ours is trained/evaluated by the proposed algorithm. Since involving the proposed adaptive residual layer has little impact on the model complexity\footnote{The number of parameters/FLOPs increases from 7.761M/54.83G to 7.762M/55.17G after involving the proposed adaptive residual layer.}, we claim that they are the same architecture/backbone for consistency.
The evaluation results on SID~\cite{chen2018learning} are reported in Table \ref{tab:sid} and the results on ELD~\cite{wei2020physics} are in Table \ref{tab:eld}. As we can see in these tables, the deep learning-based methods tend to achieve better performance than the non-deep methods even if trained without using paired-real data. Besides, the proposed method achieves better performance than P+G \cite{foi2008practical} under the same noise model and backbone network. Some visual results are shown in Fig. \ref{fig:vis_res}. 
\begin{figure}[htbp]
    \centering
    \begin{subfigure}{0.48\linewidth}
    \includegraphics[width=\linewidth]{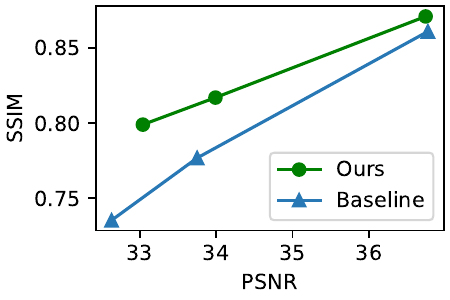}
    \vspace{-0.35cm}
    \caption{The performance of models under amplifying ratios of $[100, 250, 300]$.}
    \end{subfigure}
    \hspace{0.1cm}
    \begin{subfigure}{0.48\linewidth}
    \includegraphics[width=\linewidth]{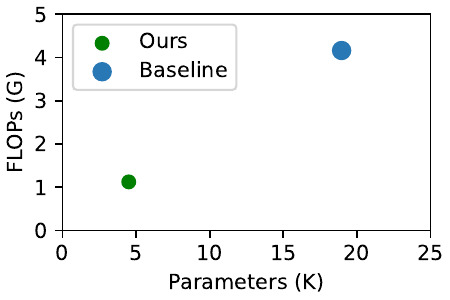}
    \vspace{-0.55cm}
    \caption{Comparisons of models in the number of parameters, FLOPs, and inference time.}
    \end{subfigure}
    \vspace{-0.7cm}
    \caption{The comparison on SID~\cite{chen2018learning} dataset between a small model (ours) with the proposed method, and a baseline larger model. The proposed method can achieve better performance using around 25\% of parameters and FLOPs of the larger model. Even taken iterations into consideration, the inference time (represented as the point size in (b)) of the proposed method is still shorter than the larger model.}
    \label{fig:perf_vs_size}
    \vspace{-0.1cm}
\end{figure}

\begin{table}[htbp]
    \centering
    \scalebox{0.85}{
    \begin{tabular}{lcccccc}
    \toprule
    & $\times 100$ & $\times 250$ & $\times 300$ \\
    Model & PSNR / SSIM  & PSNR / SSIM & PSNR / SSIM \\
    \midrule
    BM3D~\cite{dabov2007image} & $32.92 / 0.758$ & $29.56 / 0.686$ & $28.88 / 0.674$ \\
    A-BM3D~\cite{makitalo2010optimal} & $33.79 / 0.743$ & $27.24 / 0.518$ & $26.52 / 0.558$ \\
    N2N~\cite{lehtinen2018noise2noise} & $37.42 / 0.853$ & $33.48 / 0.725$ & $32.37 / 0.686$ \\
    P+G \cite{foi2008practical, wei2020physics} & $38.31 / 0.884$ & $34.39 / 0.765$ & $33.37 / 0.730$ \\
    Ours & \textbf{38.89 / 0.902} & \textbf{36.02 / 0.832} & \textbf{35.00 / 0.808}\\
    \bottomrule
    \end{tabular}
    }
    \vspace{-0.3cm}
    \caption{Quantitative results on Sony subset of SID.}
    \label{tab:sid}
    \vspace{-0.1cm}
\end{table}

\begin{table}[]
    \centering
    \scalebox{0.85}{
    \begin{tabular}{c|c|c|ccccc}
    \toprule
        Camera & Ratio & Metrics & BM3D & N2N & P+G & Ours   \\
    \midrule
    \multirow{4}{*}{\shortstack{Sony\\A7S2}} & \multirow{2}{*}{$\times 100$} & PSNR & 37.69 & 41.63 & \second{42.46} & \best{43.29}  \\
    & & SSIM & 0.803 & 0.856 & \second{0.889} & \best{0.929} \\
    \cline{2-7}
    & \multirow{2}{*}{$\times 200$} & PSNR & 34.06 & 37.98 & \second{38.88} & \best{40.39} \\
    & & SSIM & 0.696 & 0.775 & \second{0.812} & \best{0.873} \\
    \hline
    \multirow{4}{*}{\shortstack{Nikon\\D850}} & \multirow{2}{*}{$\times 100$} & PSNR & 33.97 & 40.47 & \second{40.29} & \best{40.89}  \\
    & & SSIM & 0.725 & 0.848 & \second{0.845} & \best{0.897} \\
    \cline{2-7}
    & \multirow{2}{*}{$\times 200$} & PSNR & 31.36 & 37.98 & \second{37.26} & \best{38.51} \\
    & & SSIM & 0.618 & 0.820 & \second{0.786} & \best{0.856}\\
    \hline
    \multirow{4}{*}{\shortstack{Canon\\EOS70D}} & \multirow{2}{*}{$\times 100$} & PSNR & 30.79 & 38.21 & \second{40.94} & \best{40.99} \\
    & & SSIM & 0.589 & 0.826 & \second{0.934} & \best{0.944}\\
    \cline{2-7}
    & \multirow{2}{*}{$\times 200$} & PSNR & 28.06 & 34.33 & \second{37.64} & \best{37.90}\\
    & & SSIM & 0.540 & 0.704 & \second{0.873} & \best{0.874}\\ 
    \hline
    \multirow{4}{*}{\shortstack{Canon\\EOS700D}} & \multirow{2}{*}{$\times 100$} & PSNR & 29.70 & 38.29 & \second{40.08} & \best{40.19} \\
    & & SSIM & 0.556 & 0.859 & \second{0.897} & \best{0.918}\\
    \cline{2-7}
    & \multirow{2}{*}{$\times 200$} & PSNR & 27.52 & 34.94 & \best{37.86} & \second{37.71} \\
    & & SSIM & 0.537 & 0.766 & \best{0.879} & \second{0.878} \\
    \bottomrule
    \end{tabular}
    }
    \vspace{-0.1cm}
    \caption{The quantitative results on ELD~\cite{wei2020physics} dataset.}
    \label{tab:eld}
\end{table}

\subsection{The results with different noise models}
\label{sec:noise_model}
To explore whether the proposed strategy is compatible with different modeling of the exposure process, we further evaluate the performance of models trained on synthetic data synthesized by SOTA noise models~\cite{wei2020physics, feng2022learnability} and paired real data. Specifically, compared with P-G noise model~\cite{foi2008practical}, \cite{wei2020physics} additionally models the row noise and further refines the modeling of signal-independent noise. Most recently, \cite{feng2022learnability} collects a set of dark frames to correct the dark shading. By subtracting the dark shading in the preprocessing pipeline, the performance is further improved.
%
As shown in Table \ref{tab:noise_model}, the more accurate noise models we use, the better performance we achieved. The models based on SOTA noise models even achieved slightly better performance than that trained on the real paired data. 
%
%
By training models using the proposed method, we achieve consistent improvements in performance across all models. the results serve as a strong indication of the effectiveness and generality of the proposed model.

\begin{table}[]
    \centering
    \scalebox{0.9}{
    \begin{tabular}{clcc}
    \toprule
       & \multirow{2}{*}{Model} & Baseline & w/ Ours\\
       &  & PSNR / SSIM  & PSNR / SSIM \\
    \midrule
    \multirow{4}{*}{$\times100$} & P+G~\cite{foi2008practical} & 38.31 / 0.884 & \textbf{38.89} / \textbf{0.902} \\
    &    Paired data & 38.60 / 0.912 & \textbf{38.98} / \textbf{0.915}\\
    &    ELD~\cite{wei2020physics} & 39.27 / 0.914 & \textbf{39.37} / \textbf{0.917}\\
    & PMN~\cite{feng2022learnability} & 39.77 / 0.919 & \textbf{39.80} / \textbf{0.920} \\
    \hline
    \multirow{4}{*}{$\times250$} & P+G~\cite{foi2008practical} & 34.39 / 0.765 & \textbf{36.02} / \textbf{0.832} \\
    &    Paired data & 37.08 / 0.886 & \textbf{37.45} / \textbf{0.895}\\
    &    ELD~\cite{wei2020physics} & 37.13 / 0.883 & \textbf{37.47} / \textbf{0.889}\\
    & PMN~\cite{feng2022learnability} & 37.68 / 0.892 & \textbf{37.90} / \textbf{0.896} \\
    \hline
    \multirow{4}{*}{$\times300$} & P+G~\cite{foi2008practical} & 33.37 / 0.730 & \textbf{34.99} / \textbf{0.808} \\
    &    Paired data & 36.29 / 0.874 & \textbf{36.82} / \textbf{0.888}\\
    &    ELD~\cite{wei2020physics} & 36.30 / 0.872 & \textbf{36.78} / \textbf{0.878}\\
    & PMN~\cite{feng2022learnability} & 37.01 / 0.881 & \textbf{37.27} / \textbf{0.888} \\
    \bottomrule
    \end{tabular}}
    \vspace{-0.2cm}
    \caption{The performance of models trained with paired real data and different noise models on SID~\cite{chen2018learning}.}
    \label{tab:noise_model}
    \vspace{-0.5cm}
\end{table}

\begin{figure*}[t]
    \centering
    \captionsetup[subfigure]{justification=centering}
    \hspace{-0.1cm}
    \scalebox{1}{\begin{subfigure}{0.195\linewidth}
    \includegraphics[width=\linewidth]{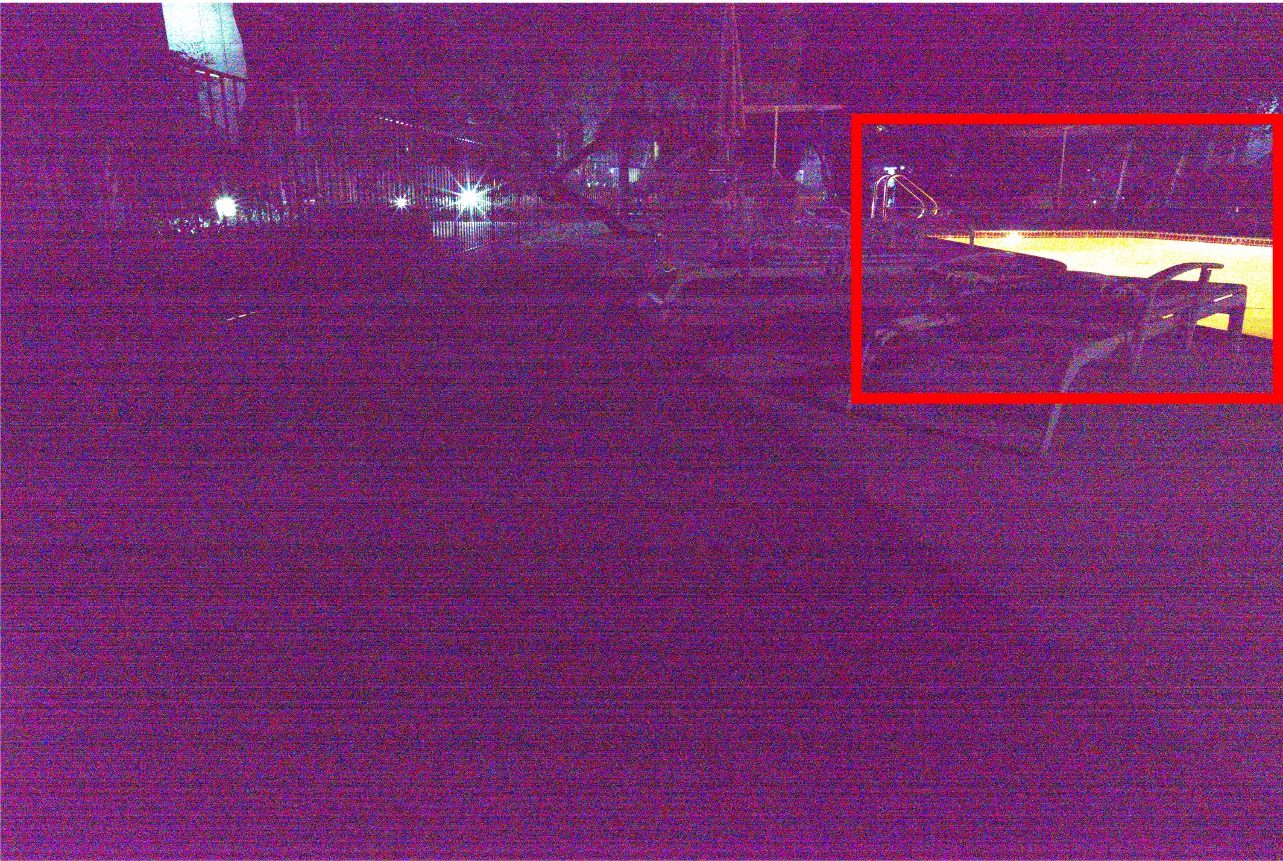}
    \caption{Low light image}
    \end{subfigure}
    \begin{subfigure}{0.195\linewidth}
    \includegraphics[width=\linewidth]{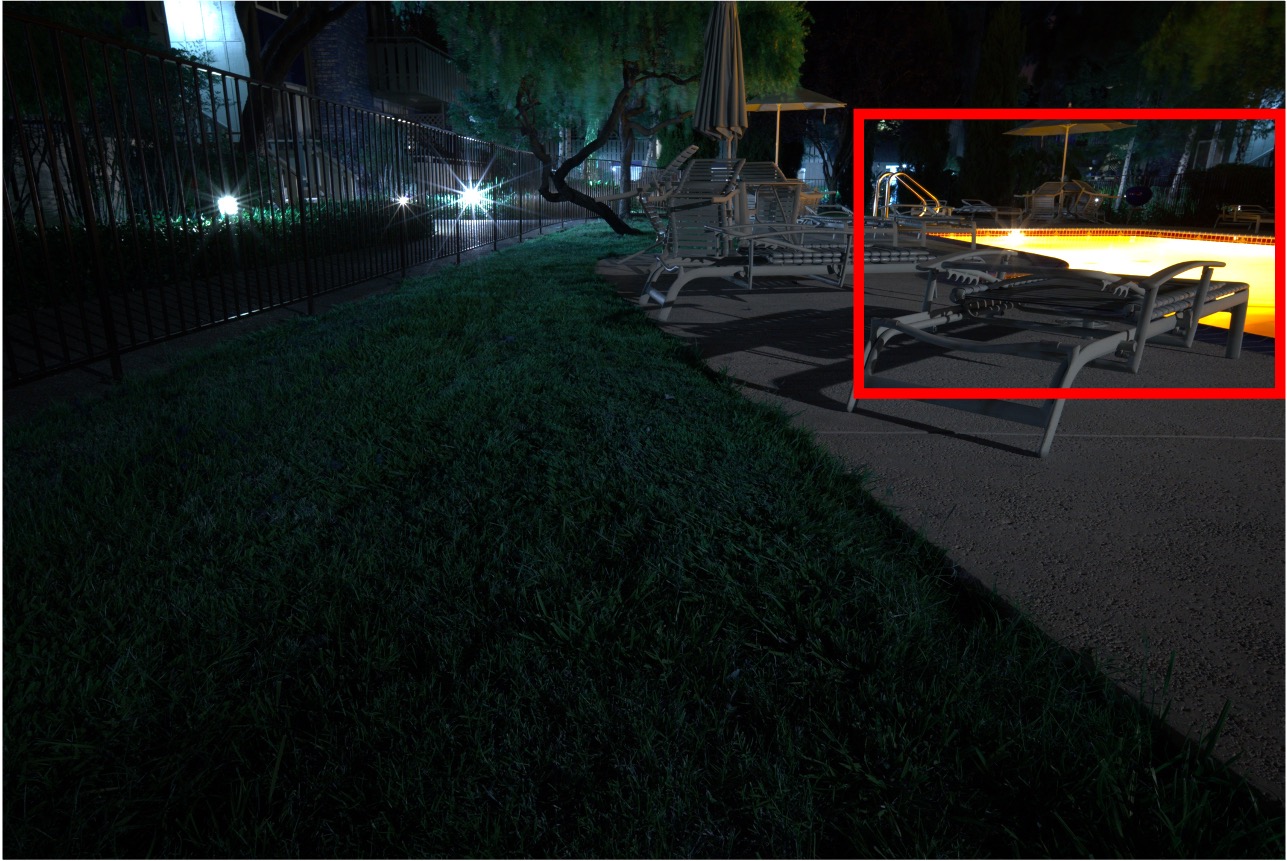}
    \caption{Reference image}
    \end{subfigure}
    \begin{subfigure}{0.195\linewidth}
    \includegraphics[width=\linewidth]{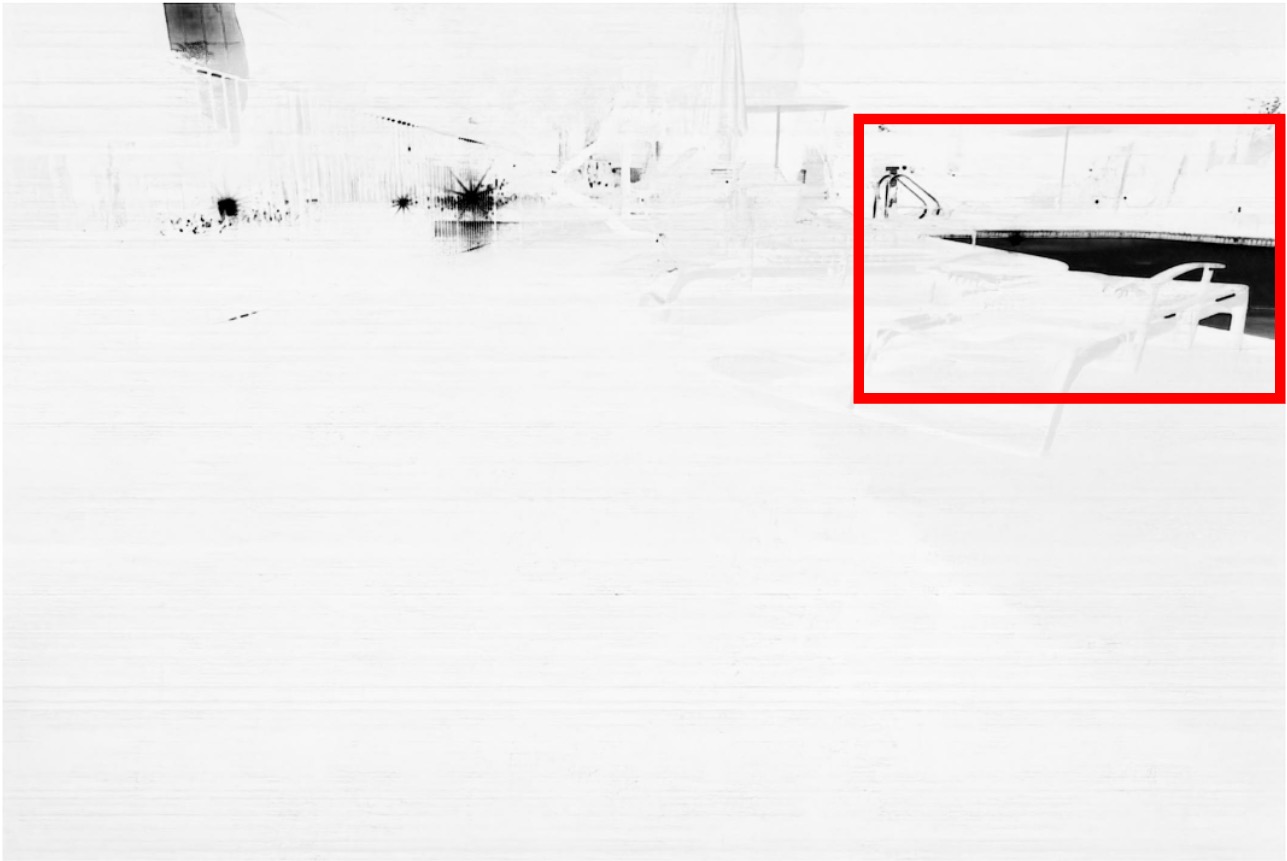}
    \caption{Adaptive mask $M$}
    \end{subfigure}
    \begin{subfigure}{0.2\linewidth}
    \includegraphics[width=\linewidth]{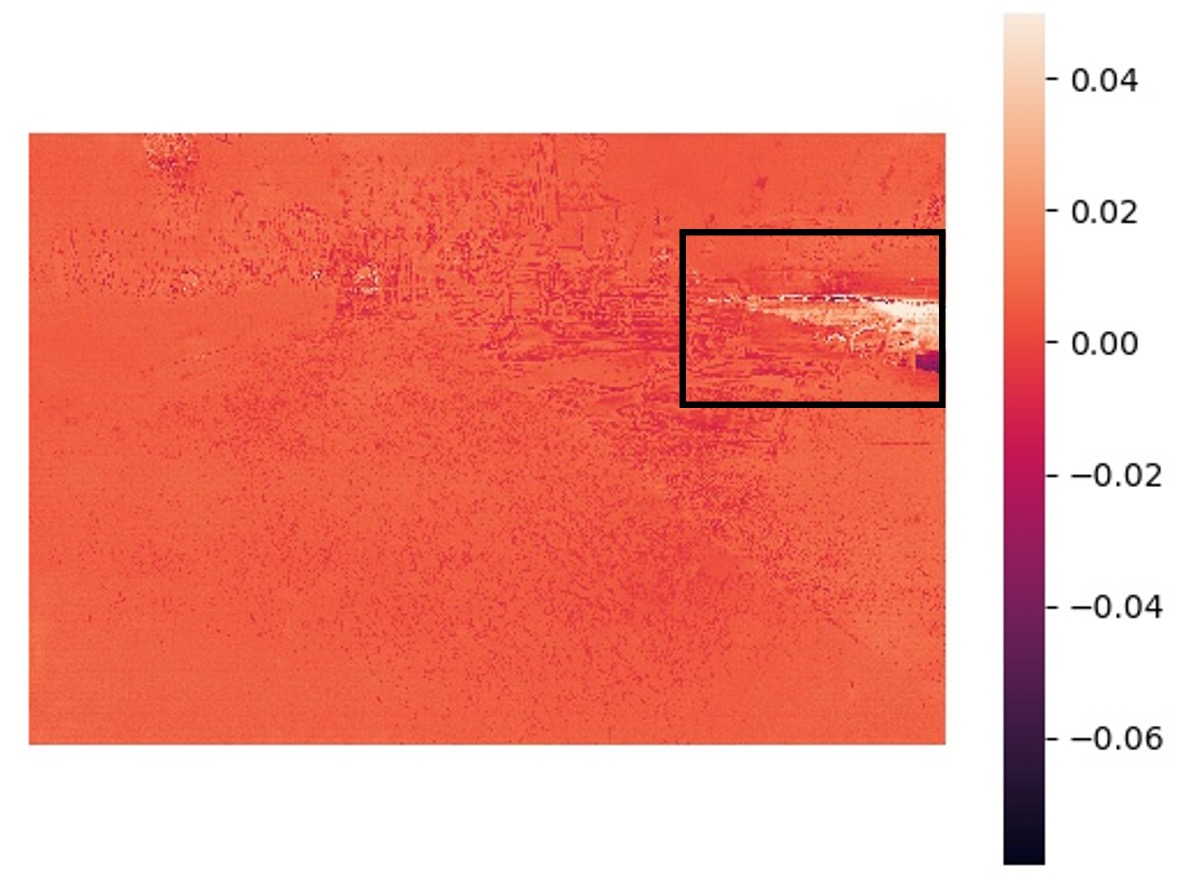}
    \caption{w/o ARL (PSNR 32.57)}
    \end{subfigure}
    \begin{subfigure}{0.2\linewidth}
    \includegraphics[width=\linewidth]{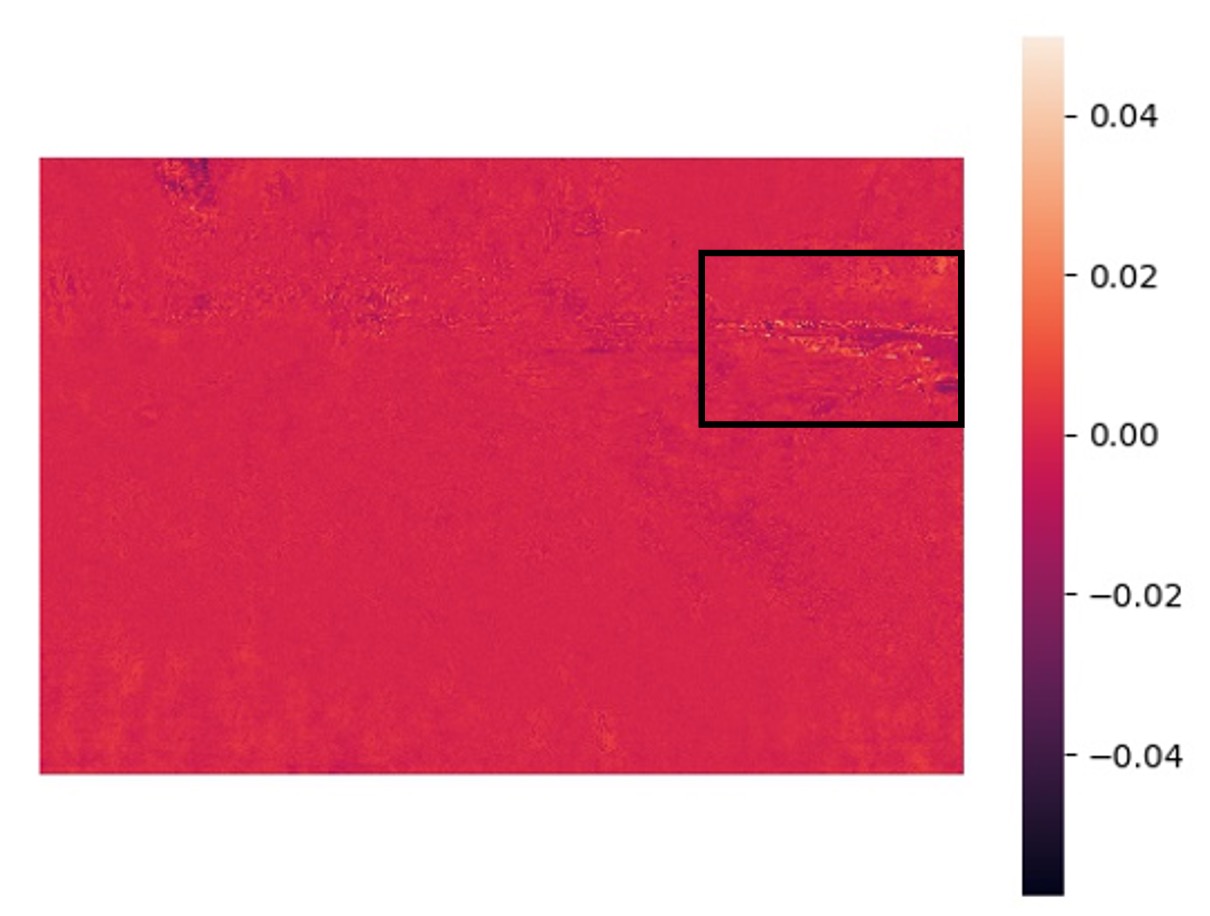}
    \caption{w/ ARL (PSNR 35.44)}
    \end{subfigure}
    \hspace{-0.2cm}}
    
    \vspace{-0.2cm}
    \caption{An illustration of the proposed adaptive residual layer (ARL). (d) and (e) are the maps of error magnitude change of models w/o and w/ ARL after iterative refinement. For the highlight areas, \eg, the area in the bounding box, the iterative refinement tends to lead to a larger value of the absolute error if w/o the proposed ARL.}
    \label{fig:adaptive}
\end{figure*}

\vspace{-0.1cm}
\subsection{The results with different backbone models}
To further explore whether the proposed method is compatible with different model sizes, we conduct experiments utilizing different backbone networks and model sizes. For the backbone networks, the widely used UNet~\cite{wei2020physics} backbone and a most recent SOTA backbone NAFNet~\cite{chen2022simple} are used for evaluation. We also evaluate NAFNet~\cite{chen2022simple} with different model sizes to explore the effect of different model capacities. The performance of different backbones and model sizes are reported in Table \ref{tab:backbone}. As shown in the table, the proposed method can stably improve the image quality of enhancement results under different backbones/model sizes. Besides, the most significant improvement is achieved for small models, which makes it possible to deploy a small model on mobile devices and introduce the proposed inference algorithm to boost performance for extreme low-light cases. The verification can be seen in Fig. \ref{fig:perf_vs_size}, in which we find that the proposed method can achieve better performance using only around $25\%$ parameters and FLOPs of the larger feedforward model. It is still faster than the larger one for inference even if we use an iteration number of $3$, and can be further sped up by using a smaller number of iteration steps or no iteration for cases that are not very dark.
\begin{table}
    \centering
    \begin{subtable}{\linewidth}
    \scalebox{0.85}{
    \begin{tabular}{clccc}
    \toprule
       & \multirow{2}{*}{\makecell[c]{Noise\\model}} & \multirow{2}{*}{Model} & Baseline & w/ Ours\\
       & &  & PSNR / SSIM  & PSNR / SSIM \\
    \midrule
    \multirow{3}{*}{$\times100$} & \multirow{3}{*}{P+G~\cite{foi2008practical}} & UNet & 38.31 / 0.884 & \textbf{38.88} / \textbf{0.901} \\
    & & NAFNet-1 & 32.37 / 0.698  & \textbf{36.74} / \textbf{0.871}\\
    & & NAFNet-2 & 39.32 / 0.909 & \textbf{39.71} / \textbf{0.917}\\
    \hline
    \multirow{3}{*}{$\times100$} & \multirow{3}{*}{ELD~\cite{wei2020physics}} & UNet & 39.27 / 0.914 & \textbf{39.37} / \textbf{0.917} \\
    & & NAFNet-1 & 35.92 / 0.848 & \textbf{36.87} / \textbf{0.879}\\
    & & NAFNet-2 & 39.52 / 0.917 & \textbf{39.75} / \textbf{0.918} \\   
    \hline\hline
        \multirow{3}{*}{$\times250$} & \multirow{3}{*}{P+G~\cite{foi2008practical}} & UNet & 34.39 / 0.765 & \textbf{36.02} / \textbf{0.832}\\
    & & NAFNet-1 & 30.33 / 0.594 & \textbf{33.99} / \textbf{0.817} \\
    & & NAFNet-2 & 36.57 / 0.849 & \textbf{37.58} / \textbf{0.886}\\
    \hline
    \multirow{3}{*}{$\times250$} & \multirow{3}{*}{ELD~\cite{wei2020physics}} & UNet & 37.13 / 0.883 & \textbf{37.47} / \textbf{0.889} \\
    & & NAFNet-1 & 32.34 / 0.726 & \textbf{34.07} / \textbf{0.821 }\\
    & & NAFNet-2 & 37.39 / 0.884 & \textbf{37.90} / \textbf{0.889} \\   
    \hline\hline
    \multirow{3}{*}{$\times300$} & \multirow{3}{*}{P+G~\cite{foi2008practical}} & UNet & 33.37 / 0.730 & \textbf{34.59} / \textbf{0.798}  \\
    & & NAFNet-1 & 29.52 / 0.549 & \textbf{33.04} / \textbf{0.794}\\
    & & NAFNet-2 & 35.83 / 0.835 & \textbf{36.90} / \textbf{0.877}\\
    \hline
    \multirow{3}{*}{$\times300$} & \multirow{3}{*}{ELD~\cite{wei2020physics}} & UNet & 36.30 / 0.872 & \textbf{36.78} / \textbf{0.878}\\
    & & NAFNet-1 & 31.05 / 0.668 & \textbf{33.12} / \textbf{0.800}\\
    & & NAFNet-2 & 36.53 / 0.871 & \textbf{37.23} / \textbf{0.884} \\   
    \bottomrule
    \end{tabular}}
    \vspace{-2mm}
    \caption{Performance with different models.}
    \end{subtable}

    \begin{subtable}{\linewidth}
        \centering
        \scalebox{0.85}{\begin{tabular}{cccc}
        \toprule
         Model & Parameters & FLOPs & Inference time \\
        \midrule
         UNet & 7.762M & 55.17G & 0.1243s \\
         NAFNet-1 & 4.697K & 1.124G & 0.0468s \\
         NAFNet-2 & 6.871M & 15.32G & 0.3514s \\
        \bottomrule
        \end{tabular}}
        \vspace{-2mm}
        \caption{Computational cost of each model.}
        \vspace{-0.4cm}
    \end{subtable} 
    \vspace{-0.3cm}
    \caption{Performance of models w/ and w/o the proposed method under different noise models and backbones.}
    \vspace{-0.1cm}
    \label{tab:backbone}
\end{table}
    
\begin{figure}[htbp]
    \centering
    \begin{subfigure}{0.48\linewidth}
    \includegraphics[width=\linewidth, trim=2 1 1 2, clip]{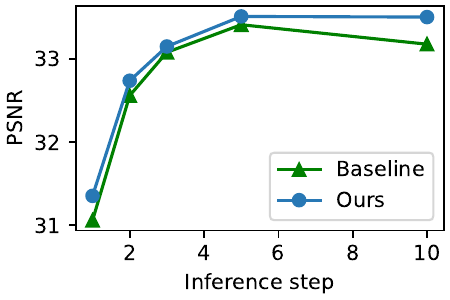}
    \end{subfigure}
    \begin{subfigure}{0.48\linewidth}
    \includegraphics[width=\linewidth, trim=2 1 1 2, clip]{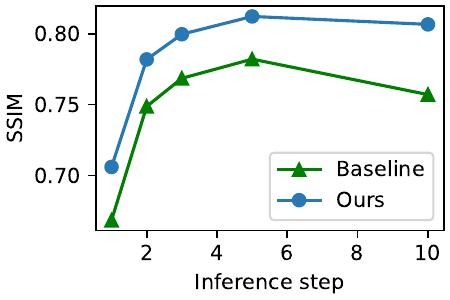}
    \end{subfigure}
    \vspace{-0.1cm}
    \caption{The performance of models with different numbers of iterations for $\F$. \textit{Baseline} represents the model w/o the proposed training paradigm and improved architecture. All the models are trained based on the NAFNet-1 backbone in Table \ref{tab:noise_model} (b) and ELD~\cite{wei2020physics} noise model.}
    \label{fig:ablation_iter}
\end{figure}
    
\vspace{-0.05cm}
\subsection{Generalization ability}
\vspace{-0.05cm}
One challenge of the existing methods is that better performance on benchmarking datasets may lead to worse out-of-distribution (O.O.D) performance. To evaluate the generalization ability of the proposed model, we train models on the $\times 100$ task of SID dataset \cite{chen2018learning} and evaluates their performance on the $\times 250$ and $\times 300$ tasks. The SOTA noise model and network architecture are utilized, \ie, PMN \cite{feng2022learnability} and NAFNet~\cite{chen2022simple}. The results are reported in Table \ref{tab:generalization}. As shown in the table, the proposed method can alleviate the performance degradation caused by the domain gap. We conjecture the reason is that even if a single step of denoising is not accurate enough, by involving slightly more denoising steps, the performance gap can be alleviated.   

\begin{table}[tbp]
    \centering
    
    \scalebox{0.9}{
    \begin{tabular}{cccc}
    \toprule
       & Method &  PSNR & SSIM\\
    \midrule
    \multirow{2}{*}{$\times 250$} & Baseline & 37.37 & 0.8741  \\
    & Ours & \textbf{37.73} & \textbf{0.8846}\\
    \midrule
    \multirow{2}{*}{$\times 300$} & Baseline & 36.31 & 0.8512 \\
    & Ours & \textbf{36.83} & \textbf{0.8650} \\
    \bottomrule
    \end{tabular}
    }
    \vspace{-0.1cm}
    \caption{The generalization ability of methods on O.O.D tasks. The models are trained on the $\times 100$ task and evaluated on $\times 250$ and $\times 300$ tasks. All the models use the same SOTA noise model PMN~\cite{feng2022learnability} and backbone NAFNet-2~\cite{chen2022simple}. \textit{Baseline} is not equipped with the improved architecture and the proposed training paradigm.}
    \label{tab:generalization}
    \vspace{-0.3cm}
\end{table}

\vspace{-0.1cm}
\subsection{Ablation study}
\vspace{-0.1cm}
\para{The impact of different inference steps.} Similar to previous works that involve progressive refinement, the number of inference steps is a key hyper-parameter that we need to set manually. Therefore, we evaluate the effect of different inference steps and the results can be seen in Fig. \ref{fig:ablation_iter}. As shown in the figure, in the first few steps, the performance of all the models increases monotonously. However, the baseline method has an obvious performance degradation when the number of inference steps increases to $10$, due to the mismatch between the inference distribution and the training one. Benefiting from the better matching of the training and testing distribution and the improved architecture, the proposed method has better initial results and performs relatively stable at a relatively large inference step. It is worth noting that we evaluate the metrics on $\hat{X}_{ref}$ of each step instead of $X_{t}$, which are more accurate to evaluate the effectiveness of the proposed method.

\noindent\textbf{Adaptive residual layer.} To better understand the role of the proposed adaptive residual layer, an example is provided in Fig. \ref{fig:adaptive}. As we can see in the figure, the highlight areas of the low-light image and reference image are almost the same after multiplying the amplifying ratio to the low-light image \cite{huang2022towards}. Progressively refining the highlight areas may make their values deviate from the real values due to the inductive bias of neural networks if we directly predict a clean image from a noisy one. By introducing the proposed adaptive residual layer, the model relies more on the results by residual-based denoising, \ie, the second term in Eq. \ref{eq:ARL}, which tends to predict noises with zero mean so that the intensity level is less affected in the highlight areas. As shown in the figure, the results after iteration without the proposed ARL degrade the fidelity in highlights while the proposed method greatly solves this problem.

\vspace{-0.1cm}
\section{Conclusion}
\vspace{-0.1cm}
In this paper, we propose a novel strategy for raw-image enhancement. Specifically, we propose to utilize a shared-weight network to simulate the physics-based exposure process by minimizing their KL divergence in an iterative end-to-end manner. An adaptive residual layer is further proposed to alleviate the fidelity deterioration caused by the iterative refinement in the highlight areas. We evaluate the effectiveness of the proposed algorithm on two benchmarks and the results demonstrate that the proposed method can stably improve the performance combined with real-paired data, different noise models, and different backbones. Besides, the proposed method also achieves better generalization ability in unseen amplifying ratios.



{\small
\bibliographystyle{ieee_fullname}
\bibliography{egbib}

\begin{thebibliography}{10}\itemsep=-1pt

\bibitem{chen2018learning}
Chen Chen, Qifeng Chen, Jia Xu, and Vladlen Koltun.
\newblock Learning to see in the dark.
\newblock In {\em Proceedings of the IEEE Conference on Computer Vision and
  Pattern Recognition}, pages 3291--3300, 2018.

\bibitem{chen2022simple}
Liangyu Chen, Xiaojie Chu, Xiangyu Zhang, and Jian Sun.
\newblock Simple baselines for image restoration.
\newblock In {\em Computer Vision--ECCV 2022: 17th European Conference, October
  23--27, 2022, Proceedings, Part VII}, pages 17--33, 2022.

\bibitem{chung2022improving}
Hyungjin Chung, Byeongsu Sim, Dohoon Ryu, and Jong~Chul Ye.
\newblock Improving diffusion models for inverse problems using manifold
  constraints.
\newblock {\em arXiv preprint arXiv:2206.00941}, 2022.

\bibitem{dabov2007image}
Kostadin Dabov, Alessandro Foi, Vladimir Katkovnik, and Karen Egiazarian.
\newblock Image denoising by sparse 3-d transform-domain collaborative
  filtering.
\newblock {\em IEEE Transactions on Image Processing}, 16(8):2080--2095, 2007.

\bibitem{dong2022abandoning}
Xingbo Dong, Wanyan Xu, Zhihui Miao, Lan Ma, Chao Zhang, Jiewen Yang, Zhe Jin,
  Andrew Beng~Jin Teoh, and Jiajun Shen.
\newblock Abandoning the bayer-filter to see in the dark.
\newblock In {\em Proceedings of the IEEE/CVF Conference on Computer Vision and
  Pattern Recognition}, pages 17431--17440, 2022.

\bibitem{feng2022learnability}
Hansen Feng, Lizhi Wang, Yuzhi Wang, and Hua Huang.
\newblock Learnability enhancement for low-light raw denoising: Where paired
  real data meets noise modeling.
\newblock In {\em Proceedings of the 30th ACM International Conference on
  Multimedia}, pages 1436--1444, 2022.

\bibitem{foi2008practical}
Alessandro Foi, Mejdi Trimeche, Vladimir Katkovnik, and Karen Egiazarian.
\newblock Practical poissonian-gaussian noise modeling and fitting for
  single-image raw-data.
\newblock {\em IEEE transactions on image processing}, 17(10):1737--1754, 2008.

\bibitem{graikosdiffusion}
Alexandros Graikos, Nikolay Malkin, Nebojsa Jojic, and Dimitris Samaras.
\newblock Diffusion models as plug-and-play priors.
\newblock In {\em Advances in Neural Information Processing Systems}.

\bibitem{guo2022enhancing}
Lanqing Guo, Renjie Wan, Wenhan Yang, Alex Kot, and Bihan Wen.
\newblock Enhancing low-light images in real world via cross-image
  disentanglement.
\newblock {\em arXiv preprint arXiv:2201.03145}, 2022.

\bibitem{guo2022shadowdiffusion}
Lanqing Guo, Chong Wang, Wenhan Yang, Siyu Huang, Yufei Wang, Hanspeter
  Pfister, and Bihan Wen.
\newblock Shadowdiffusion: When degradation prior meets diffusion model for
  shadow removal.
\newblock 2023.

\bibitem{guo2022low}
Xiaojie Guo and Qiming Hu.
\newblock Low-light image enhancement via breaking down the darkness.
\newblock {\em International Journal of Computer Vision}, pages 1--19, 2022.

\bibitem{ho2020denoising}
Jonathan Ho, Ajay Jain, and Pieter Abbeel.
\newblock Denoising diffusion probabilistic models.
\newblock {\em Advances in Neural Information Processing Systems},
  33:6840--6851, 2020.

\bibitem{huang2022towards}
Haofeng Huang, Wenhan Yang, Yueyu Hu, Jiaying Liu, and Ling-Yu Duan.
\newblock Towards low light enhancement with raw images.
\newblock {\em IEEE Transactions on Image Processing}, 31:1391--1405, 2022.

\bibitem{huang2022deep}
Jie Huang, Yajing Liu, Feng Zhao, Keyu Yan, Jinghao Zhang, Yukun Huang, Man
  Zhou, and Zhiwei Xiong.
\newblock Deep fourier-based exposure correction network with spatial-frequency
  interaction.
\newblock In {\em Computer Vision--ECCV 2022: 17th European Conference, 2022,
  Proceedings, Part XIX}, pages 163--180. Springer, 2022.

\bibitem{jiniccv23led}
Xin Jin, Jia-Wen Xiao, Ling-Hao Han, Chunle Guo, Ruixun Zhang, Xialei Liu, and
  Chongyi Li.
\newblock Lighting every darkness in two pairs: A calibration-free pipeline for
  raw denoising.
\newblock 2023.

\bibitem{jin2023enhancing}
Yeying Jin, Beibei Lin, Wending Yan, Wei Ye, Yuan Yuan, and Robby~T. Tan.
\newblock Enhancing visibility in nighttime haze images using guided apsf and
  gradient adaptive convolution, 2023.

\bibitem{jin2022unsupervised}
Yeying Jin, Wenhan Yang, and Robby~T Tan.
\newblock Unsupervised night image enhancement: When layer decomposition meets
  light-effects suppression.
\newblock In {\em Computer Vision--ECCV 2022: 17th European Conference, Tel
  Aviv, Israel, October 23--27, 2022, Proceedings, Part XXXVII}, pages
  404--421. Springer, 2022.

\bibitem{kawardenoising}
Bahjat Kawar, Michael Elad, Stefano Ermon, and Jiaming Song.
\newblock Denoising diffusion restoration models.
\newblock In {\em Advances in Neural Information Processing Systems}.

\bibitem{lehtinen2018noise2noise}
Jaakko Lehtinen, Jacob Munkberg, Jon Hasselgren, Samuli Laine, Tero Karras,
  Miika Aittala, and Timo Aila.
\newblock Noise2noise: Learning image restoration without clean data.
\newblock {\em arXiv preprint arXiv:1803.04189}, 2018.

\bibitem{li2021low}
Chongyi Li, Chunle Guo, Linghao Han, Jun Jiang, Ming-Ming Cheng, Jinwei Gu, and
  Chen~Change Loy.
\newblock Low-light image and video enhancement using deep learning: A survey.
\newblock {\em IEEE transactions on pattern analysis and machine intelligence},
  44(12):9396--9416, 2021.

\bibitem{liu2021benchmarking}
Jiaying Liu, Dejia Xu, Wenhan Yang, Minhao Fan, and Haofeng Huang.
\newblock Benchmarking low-light image enhancement and beyond.
\newblock {\em International Journal of Computer Vision}, 129(4):1153--1184,
  2021.

\bibitem{lore2017llnet}
Kin~Gwn Lore et~al.
\newblock {LLNet}: A deep autoencoder approach to natural low-light image
  enhancement.
\newblock {\em Pattern Recognition}, 61:650--662, 2017.

\bibitem{lugmayr2022repaint}
Andreas Lugmayr, Martin Danelljan, Andres Romero, Fisher Yu, Radu Timofte, and
  Luc Van~Gool.
\newblock Repaint: Inpainting using denoising diffusion probabilistic models.
\newblock In {\em Proceedings of the IEEE/CVF Conference on Computer Vision and
  Pattern Recognition}, pages 11461--11471, 2022.

\bibitem{lugmayr2020srflow}
Andreas Lugmayr, Martin Danelljan, Luc Van~Gool, and Radu Timofte.
\newblock Srflow: Learning the super-resolution space with normalizing flow.
\newblock In {\em European Conference on Computer Vision}, pages 715--732.
  Springer, 2020.

\bibitem{lv2018mbllen}
Feifan Lv, Feng Lu, Jianhua Wu, and Chongsoon Lim.
\newblock Mbllen: Low-light image/video enhancement using cnns.
\newblock In {\em BMVC}, page 220, 2018.

\bibitem{ma2022accelerating}
Hengyuan Ma, Li Zhang, Xiatian Zhu, and Jianfeng Feng.
\newblock Accelerating score-based generative models with preconditioned
  diffusion sampling.
\newblock In {\em Computer Vision--ECCV 2022: 17th European Conference, Tel
  Aviv, Israel, October 23--27, 2022, Proceedings, Part XXIII}, pages 1--16.
  Springer, 2022.

\bibitem{makitalo2010optimal}
Markku Makitalo and Alessandro Foi.
\newblock Optimal inversion of the anscombe transformation in low-count poisson
  image denoising.
\newblock {\em IEEE transactions on Image Processing}, 20(1):99--109, 2010.

\bibitem{pokledeep}
Ashwini Pokle, Zhengyang Geng, and J~Zico Kolter.
\newblock Deep equilibrium approaches to diffusion models.
\newblock In {\em Advances in Neural Information Processing Systems}.

\bibitem{punnappurath2022day}
Abhijith Punnappurath, Abdullah Abuolaim, Abdelrahman Abdelhamed, Alex
  Levinshtein, and Michael~S Brown.
\newblock Day-to-night image synthesis for training nighttime neural isps.
\newblock In {\em Proceedings of the IEEE/CVF Conference on Computer Vision and
  Pattern Recognition}, pages 10769--10778, 2022.

\bibitem{ren2019low}
Wenqi Ren, Sifei Liu, Lin Ma, Qianqian Xu, Xiangyu Xu, Xiaochun Cao, Junping
  Du, and Ming-Hsuan Yang.
\newblock Low-light image enhancement via a deep hybrid network.
\newblock {\em IEEE Transactions on Image Processing}, 28(9):4364--4375, 2019.

\bibitem{liu2021ruas}
Liu Risheng, Ma Long, Zhang Jiaao, Fan Xin, and Luo Zhongxuan.
\newblock Retinex-inspired unrolling with cooperative prior architecture search
  for low-light image enhancement.
\newblock In {\em Proceedings of the IEEE Conference on Computer Vision and
  Pattern Recognition}, 2021.

\bibitem{saharia2022palette}
Chitwan Saharia, William Chan, Huiwen Chang, Chris Lee, Jonathan Ho, Tim
  Salimans, David Fleet, and Mohammad Norouzi.
\newblock Palette: Image-to-image diffusion models.
\newblock In {\em ACM SIGGRAPH 2022 Conference Proceedings}, pages 1--10, 2022.

\bibitem{saharia2022image}
Chitwan Saharia, Jonathan Ho, William Chan, Tim Salimans, David~J Fleet, and
  Mohammad Norouzi.
\newblock Image super-resolution via iterative refinement.
\newblock {\em IEEE Transactions on Pattern Analysis and Machine Intelligence},
  2022.

\bibitem{shen2017msr}
Liang Shen, Zihan Yue, Fan Feng, Quan Chen, Shihao Liu, and Jie Ma.
\newblock {MSR}-net: Low-light image enhancement using deep convolutional
  network.
\newblock {\em arXiv preprint arXiv:1711.02488}, 2017.

\bibitem{song2020denoising}
Jiaming Song, Chenlin Meng, and Stefano Ermon.
\newblock Denoising diffusion implicit models.
\newblock {\em arXiv preprint arXiv:2010.02502}, 2020.

\bibitem{tao2017llcnn}
Li Tao, Chuang Zhu, Guoqing Xiang, Yuan Li, Huizhu Jia, and Xiaodong Xie.
\newblock {LLCNN}: A convolutional neural network for low-light image
  enhancement.
\newblock In {\em 2017 IEEE Visual Communications and Image Processing (VCIP)},
  pages 1--4, 2017.

\bibitem{wang2021s3rp}
Chulin Wang, Kyongmin Yeo, Xiao Jin, Andres Codas, Levente~J Klein, and Bruce
  Elmegreen.
\newblock S3rp: Self-supervised super-resolution and prediction for
  advection-diffusion process.
\newblock {\em arXiv preprint arXiv:2111.04639}, 2021.

\bibitem{wang2022ultra}
Tao Wang, Kaihao Zhang, Tianrun Shen, Wenhan Luo, Bjorn Stenger, and Tong Lu.
\newblock Ultra-high-definition low-light image enhancement: A benchmark and
  transformer-based method.
\newblock {\em arXiv preprint arXiv:2212.11548}, 2022.

\bibitem{wang2019progressive}
Yang Wang, Yang Cao, Zheng-Jun Zha, Jing Zhang, Zhiwei Xiong, Wei Zhang, and
  Feng Wu.
\newblock Progressive retinex: Mutually reinforced illumination-noise
  perception network for low-light image enhancement.
\newblock In {\em Proceedings of the 27th ACM International Conference on
  Multimedia}, pages 2015--2023, 2019.

\bibitem{wang2021low}
Yufei Wang, Renjie Wan, Wenhan Yang, Haoliang Li, Lap-Pui Chau, and Alex Kot.
\newblock Low-light image enhancement with normalizing flow.
\newblock In {\em Proceedings of the AAAI Conference on Artificial
  Intelligence}, volume~36, pages 2604--2612, 2022.

\bibitem{wang2023beyond}
Yufei Wang, Yi Yu, Wenhan Yang, Lanqing Guo, Lap-Pui Chau, Alex~C Kot, and
  Bihan Wen.
\newblock Beyond learned metadata-based raw image reconstruction.
\newblock {\em arXiv preprint arXiv:2306.12058}, 2023.

\bibitem{wang2023raw}
Yufei Wang, Yi Yu, Wenhan Yang, Lanqing Guo, Lap-Pui Chau, Alex~C Kot, and
  Bihan Wen.
\newblock Raw image reconstruction with learned compact metadata.
\newblock In {\em Proceedings of the IEEE/CVF Conference on Computer Vision and
  Pattern Recognition}, pages 18206--18215, 2023.

\bibitem{wei2018deep}
Chen Wei, Wenjing Wang, Wenhan Yang, and Jiaying Liu.
\newblock Deep retinex decomposition for low-light enhancement.
\newblock {\em arXiv preprint arXiv:1808.04560}, 2018.

\bibitem{wei2020physics}
Kaixuan Wei, Ying Fu, Jiaolong Yang, and Hua Huang.
\newblock A physics-based noise formation model for extreme low-light raw
  denoising.
\newblock In {\em Proceedings of the IEEE/CVF Conference on Computer Vision and
  Pattern Recognition}, pages 2758--2767, 2020.

\bibitem{whang2022deblurring}
Jay Whang, Mauricio Delbracio, Hossein Talebi, Chitwan Saharia, Alexandros~G
  Dimakis, and Peyman Milanfar.
\newblock Deblurring via stochastic refinement.
\newblock In {\em Proceedings of the IEEE/CVF Conference on Computer Vision and
  Pattern Recognition}, pages 16293--16303, 2022.

\bibitem{wu2022uretinex}
Wenhui Wu, Jian Weng, Pingping Zhang, Xu Wang, Wenhan Yang, and Jianmin Jiang.
\newblock Uretinex-net: Retinex-based deep unfolding network for low-light
  image enhancement.
\newblock In {\em Proceedings of the IEEE/CVF Conference on Computer Vision and
  Pattern Recognition}, pages 5901--5910, 2022.

\bibitem{xu2022snr}
Xiaogang Xu, Ruixing Wang, Chi-Wing Fu, and Jiaya Jia.
\newblock Snr-aware low-light image enhancement.
\newblock In {\em Proceedings of the IEEE/CVF Conference on Computer Vision and
  Pattern Recognition}, pages 17714--17724, 2022.

\bibitem{zhang2022gddim}
Qinsheng Zhang, Molei Tao, and Yongxin Chen.
\newblock gddim: Generalized denoising diffusion implicit models.
\newblock {\em arXiv preprint arXiv:2206.05564}, 2022.

\bibitem{zhang2021rellie}
Rongkai Zhang, Lanqing Guo, Siyu Huang, and Bihan Wen.
\newblock Rellie: Deep reinforcement learning for customized low-light image
  enhancement.
\newblock In {\em Proceedings of the 29th ACM international conference on
  multimedia}, pages 2429--2437, 2021.

\bibitem{zhang2021rethinking}
Yi Zhang, Hongwei Qin, Xiaogang Wang, and Hongsheng Li.
\newblock Rethinking noise synthesis and modeling in raw denoising.
\newblock In {\em Proceedings of the IEEE/CVF International Conference on
  Computer Vision}, pages 4593--4601, 2021.

\bibitem{zheng2021adaptive}
Chuanjun Zheng, Daming Shi, and Wentian Shi.
\newblock Adaptive unfolding total variation network for low-light image
  enhancement.
\newblock In {\em Proceedings of the IEEE/CVF international conference on
  computer vision}, pages 4439--4448, 2021.

\end{thebibliography}


\begin{thebibliography}{1}\itemsep=-1pt

\bibitem{chen2018learning}
Chen Chen, Qifeng Chen, Jia Xu, and Vladlen Koltun.
\newblock Learning to see in the dark.
\newblock In {\em Proceedings of the IEEE Conference on Computer Vision and Pattern Recognition}, pages 3291--3300, 2018.

\bibitem{feng2022learnability}
Hansen Feng, Lizhi Wang, Yuzhi Wang, and Hua Huang.
\newblock Learnability enhancement for low-light raw denoising: Where paired real data meets noise modeling.
\newblock In {\em Proceedings of the 30th ACM International Conference on Multimedia}, pages 1436--1444, 2022.

\bibitem{saharia2022image}
Chitwan Saharia, Jonathan Ho, William Chan, Tim Salimans, David~J Fleet, and Mohammad Norouzi.
\newblock Image super-resolution via iterative refinement.
\newblock {\em IEEE Transactions on Pattern Analysis and Machine Intelligence}, 2022.

\bibitem{song2020denoising}
Jiaming Song, Chenlin Meng, and Stefano Ermon.
\newblock Denoising diffusion implicit models.
\newblock {\em arXiv preprint arXiv:2010.02502}, 2020.

\bibitem{wei2020physics}
Kaixuan Wei, Ying Fu, Jiaolong Yang, and Hua Huang.
\newblock A physics-based noise formation model for extreme low-light raw denoising.
\newblock In {\em Proceedings of the IEEE/CVF Conference on Computer Vision and Pattern Recognition}, pages 2758--2767, 2020.

\end{thebibliography}
}

\appendix

\section{Detail of proofs}
The detailed proof of Eq. (7) in the main paper is as follows where Jensen's inequality is used
\begin{equation*}
\small
\begin{aligned}
    &\D{\p(X_{0:T})}{q(X_{0:T})} \\
    &= \E{\p(X_{0:T})}{\log \frac{\p(X_{0:T})}{q(X_{0:T})}} \\
    &= \E{\p(X_{0:T})}{\log \frac{\p(X_{0:T})}{\E{q(X_{ref})}{q(X_{0:T} | X_{ref})}}} \\
    & \leq \E{q(X_{ref})}{\E{\p(X_{0:T})}{\log \frac{\p(X_{0:T})}{q(X_{0:T}|X_{ref})}}} \\
    &= \E{q(X_{ref})}{\E{\p(X_{0:T})}{\log \frac{\p(X_T) \prod_{t=1}^T \p(X_{t-1}|X_{t})}{q(X_T|X_{ref}) \prod_{t=1}^T q(X_{t-1}|X_{t}, X_{ref})}}} \\
    & = \E{q(X_{ref})}{\E{\p(X_{0:T})}{\frac{\p(X_T)}{q(X_T|X_{ref})}+ \sum_{t=1}^{T} \log \frac{\p(X_{t-1}|X_{t})}{q(X_{t-1}|X_{t}, X_{ref})}}} \\
    &= \E{q(X_{ref})}{\D{\p(X_T)}{q(X_T|X_{ref})} \\ &\quad\quad + \sum_{t=1}^{T} \E{\p(X_t)}{\D{\p(X_{t-1}|X_{t})}{q(X_{t-1}|X_t,X_{ref})}}}.
\end{aligned}
\end{equation*}
Since modeling the ground-truth exposure process requires scene irradiation, we need to use a new variable $X_{ref}$ , that is, a ground-truth image ideally without noise, in our derivation. However, $X_0$ may not be noise-free, so an extra step is conducted to further remove the noise in $X_0$, resulting in $\hat{X}_{ref}=\F(X_0)$, as illustrated in algorithms presented in the main paper

For the image reconstruction loss, we aim to minimize the KL divergence between $\p(X_{t-1}|X_t)$ and $q(X_{t-1}|X_t, X_{ref})$ as follows
\begin{equation*} 
\small
\begin{split}
    &\D{\p(X_{t-1}|X_t)}{q(X_{t-1}|X_t, X_{ref})} \\
    &= \int \p(X|X_t) \log \frac{\p(X|X_t)}{q(X|X_t, X_{ref})} dX\\
    &=\int \p(X|X_t) [X \log \frac{\F(X_t)}{X_{t-1}} + X_{t-1} - \F(X_t)] dX_{t-1} \\
    &= \F(X|X_t) \log \frac{\F(X_t)}{X_{t-1}} + X_{t-1} - \F(X_t).
\end{split}
\end{equation*}
We do not observe an obvious difference of the converged performance between using L1 divergence and KL divergence. The reason may be that even for the long exposure images, \eg, the reference images we used in the SID dataset, they are still not noise free. Specifically, compared to L1 loss, KL loss may be more susceptible to the influence of noise with smaller values in non-ideal situations. For simplicity, we utilize L1 loss as a substitution.
Following previous works that find using the same weighting for losses from different steps achieves slightly better performance, we keep the weight of the reconstruction loss for each step the same.

\begin{figure}[t]
    \centering
    \includegraphics[width=\linewidth]{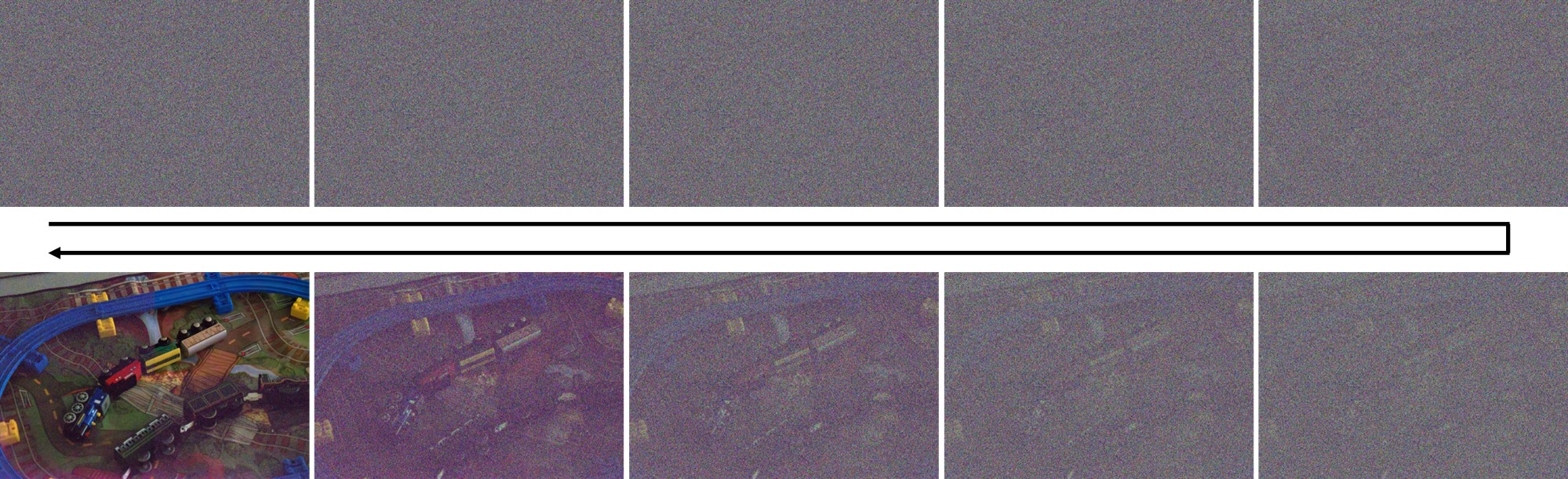}
    \caption{The reverse (denoise) process of the vanilla conditional diffusion model~\cite{saharia2022image} combined with the sampling strategy from DDIM \cite{song2020denoising}. The image restoration process starts with pure noise and gradually removes noise in it. The noise in each middle step is expected to obey Gaussian distribution so that the low-light image is not any step in it.}
    \label{fig:diffusion_x0}
\end{figure}

\section{Comparison with diffusion models}
\paragraph{Experiment setup.} To further demonstrate the superiority of the method compared with vanilla diffusion models, we compare the proposed method with a conditional diffusion model adopting the condition strategy of \cite{saharia2022image} and sampling strategy of DDIM~\cite{song2020denoising}. Specifically, similar to \cite{saharia2022image}, the low-light image is concatenated as a part of the input. 
The original backbone of \cite{song2020denoising} includes more than $100M$ parameters and $2T$ FLOPs for a single step of a $512\times512$ image patch, which is not feasible for raw images with much higher resolution, \eg, 4246$\times$2840.
Therefore, we slightly adjust the number of channels and blocks to obtain a similar model size to the commonly used backbone in raw low-light image enhancement tasks~\cite{wei2020physics,feng2022learnability}. The quantitative results are reported in Table~\ref{tab:compare_ddim}. Our method outperforms the conditional diffusion model by a large margin when using a model with similar size.

\begin{figure}[t]
    \centering
    \begin{subfigure}{0.49\linewidth}
\includegraphics[width=\linewidth]{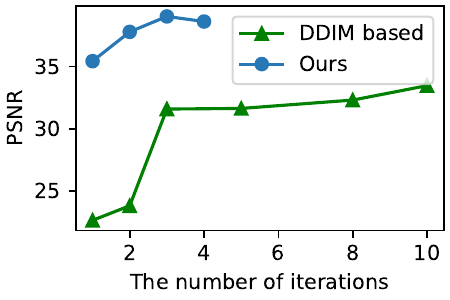}
\end{subfigure}
\begin{subfigure}{0.49\linewidth}
\includegraphics[width=\linewidth]{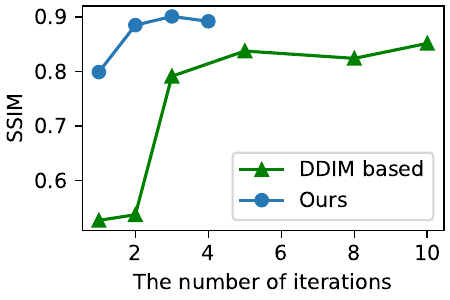}
\end{subfigure}
    \caption{The convergence of the performance under a different number of inference steps. The performances are evaluated on the $\times 100$ task of SID~\cite{chen2018learning} and models are trained with P+G noise model. The diffusion model and the proposed model have a similar number of parameters and FLOPs as shown in Table \ref{tab:compare_ddim} (b). The proposed method can get converged much faster than DDIM, leading to smaller inference overhead.}
    \label{fig:convergence}
\end{figure}

\begin{figure*}[t]
    \centering
    \includegraphics[width=\linewidth]{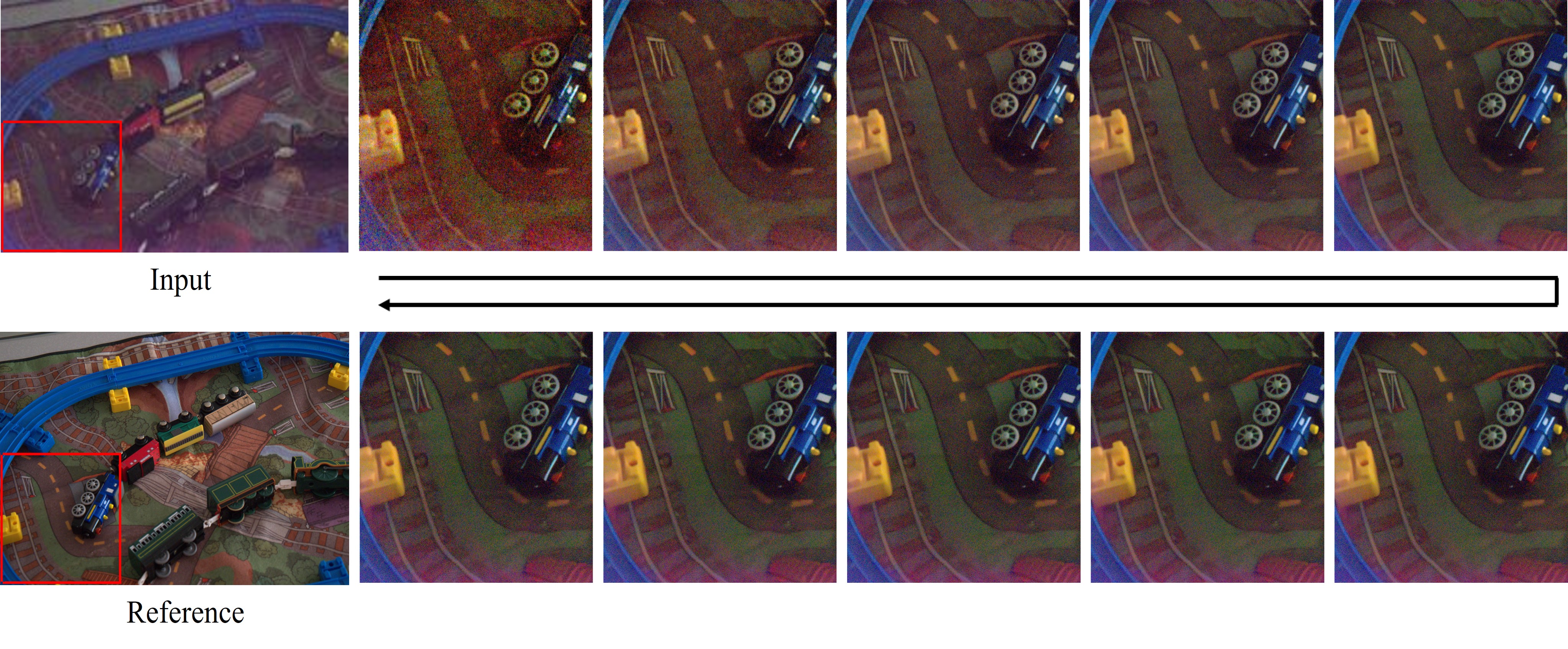}
    \caption{The diffusion process of the vanilla conditional diffusion model~\cite{saharia2022image} combined with the sampling strategy from DDIM \cite{song2020denoising}. The image restoration process starts with pure noise and gradually removes noise in it.}
    \label{fig:diffusion_xt}
\end{figure*}

\paragraph{Visualization of vanilla diffusion models.} The visualization of the vanilla conditional diffusion models is illustrated in Fig. \ref{fig:diffusion_xt} and \ref{fig:diffusion_x0}. Specifically, Fig. \ref{fig:diffusion_xt} is the inverse process that gradually removes noise from a pure noise image. For most of its middle steps, the signal-noise ratio is even lower than that of the input, causing extra inference steps. Besides, for each middle step, the noise is expected to obey the Gaussian distribution, which is different from the real noise distribution. Besides, the estimated clean image of each step is shown in Fig. \ref{fig:diffusion_x0}. From the results, we can find while the quality of the reconstructed image gradually improved, there is still intense residual noise even in images of the last few steps. We conjecture the reason is that the noise distribution of the inference process is different from the training one due to cumulative error so the noise can not be effectively removed in the last few steps. We find that the needed number of inference steps are closely related to the noise level, \eg, ISO and exposure time. An adaptive algorithm for deciding the number of inference steps may be left for future work.

\paragraph{The convergence.} 
The comparison of the convergence speed for inference between the proposed method and the vanilla one can be seen in Fig. \ref{fig:convergence}. The vanilla diffusion model needs a relatively large number (\eg, around 10 steps) of inference steps to get converged since they need to denoise from pure noise, which hinders potential applications. 
The proposed method can get converged quickly in a few steps for the cases the input is not very dark, \eg, $\times 100$ task, and the improvement mainly from the proposed training mechanism. On the contrary, for the harder cases, iterative refinement can significantly improve the performance as shown in Fig. 6 in the main paper. For the convergence of training, we find that the vanilla diffusion-based method requires a larger batch size and more epochs, \eg, we train the vanilla conditional diffusion models for $10000$ epochs with a batch size of $64$ on four RTX A5000s. While the proposed model can be well converged and trained on only one RTX A5000 with a batch size of $1$ and epoch of $300$.
 
\paragraph{Quantitative results.}
As we can see in Table~\ref{tab:compare_ddim}, the diffusion models are very sensitive to the capacity of the model size, \eg, the larger vanilla diffusion model can achieve much better performance than that of a smaller one. Besides, although the model "Diffusion-based-1" has more parameters and FLOPs than the backbone we used, it achieves much worse performance than the proposed method. The reason may be that the vanilla conditional diffusion models need more network capacity to learn the denoising of Gaussian noise with different noise levels. 

\begin{table}
    \centering
    \begin{subtable}{\linewidth}
    \centering
    \scalebox{0.95}{
    \begin{tabular}{cccc}
    \toprule
    & Model & PSNR / SSIM \\
    \midrule
    \multirow{3}{*}{$\times100$} & UNet &  \textbf{38.88} / \textbf{0.901} \\
    & Diffusion-based-1 & 33.47 / 0.851 \\
    & Diffusion-based-2 & 25.13 / 0.539\\
    \hline
        \multirow{3}{*}{$\times250$} & UNet & \textbf{36.02} / \textbf{0.832}\\
    & Diffusion-based-1 & 31.98 / 0.823 \\
    & Diffusion-based-2 & 24.79 / 0.529 \\
    \hline
    \multirow{3}{*}{$\times300$} & UNet &  \textbf{34.59} / \textbf{0.798}  \\
    & Diffusion-based-1 & 31.39 / 0.807\\
    & Diffusion-based-2 & 22.65 / 0.527 \\
    \bottomrule
    \end{tabular}}
    \caption{Performance with different models.}
    \end{subtable}

    \begin{subtable}{\linewidth}
        \centering
        \scalebox{0.95}{\begin{tabular}{cccc}
        \toprule
         Model & Parameters & FLOPs \\
        \midrule
         Ours (UNet-based) & 7.762M & 55.17G  \\
         Diffusion-based-1 & 8.619M & 367.15G \\
         Diffusion-based-2 & 1.472M & 63.07G \\
        \bottomrule
        \end{tabular}}
        \caption{Computational cost of each model.}
    \end{subtable} 
    \caption{Performance of models w/ and w/o the proposed method under different noise models and backbones on SID~\cite{chen2018learning} dataset. The models are trained on P+G noise model.}
    \label{tab:compare_ddim}
\end{table}

\section{Experiment details}
For the detailed settings of experiments, we mainly follow the settings of ELD~\cite{wei2020physics}. Specifically, we adopt the released noise parameters, and the basic noise model $P+G$ from \cite{wei2020physics}. We implement their proposed noise model $P+G^*
+r+u$ and the training on-the-fly manner by ourselves and achieve closing results with the reported in the paper. For the experiments based on PMN~\cite{feng2022learnability}, \eg, the results in Table 4 in the main paper, we follow their training strategy, \eg, the same batch size, data augmentation strategy, and learning rate. However, PMN~\cite{feng2022learnability} adapts a slightly different evaluation strategy with ELD~\cite{wei2020physics}, \eg, different evaluation set. To unify the results, we adopt the same evaluation strategy as ELD~\cite{wei2020physics} for all experiments. 
For the schedule of $\lambda_t$, we adopt the linear scheduler, \eg, the exposure time of each step for the $\times 300$ task is $[\lambda_T, 100 \cdot \lambda_T, 200 \cdot \lambda_T, 300\cdot \lambda_T]$.

\end{document}